\documentclass[journal]{IEEEtran}
\usepackage{cite}
\usepackage{amsmath,amssymb,amsfonts}
\usepackage{algorithmic}
\usepackage{graphicx}
\graphicspath{{./eps/}}
\usepackage{textcomp}
\usepackage{url}
\usepackage{booktabs}
\usepackage{multirow}
\usepackage{bm}

\def\BibTeX{{\rm B\kern-.05em{\sc i\kern-.025em b}\kern-.08em
    T\kern-.1667em\lower.7ex\hbox{E}\kern-.125emX}}

\usepackage[hidelinks]{hyperref}
\usepackage[expansion=false]{microtype}
\usepackage{etoolbox}
\makeatletter
\AtBeginDocument{%
  \patchcmd{\IEEEbiography}{plus 1fil}{plus 2pt}{}{\typeout{BIOPATCH-BEGIN-FAILED}}%
  \patchcmd{\endIEEEbiography}{plus 1fil}{plus 2pt}{}{\typeout{BIOPATCH-END-FAILED}}%
}
\makeatother
\newcommand{\orcidlink}[1]{\,\href{https://orcid.org/#1}{\textsuperscript{\scriptsize iD}}}

\begin{document}

\title{A Strong Linear Baseline for Whole-Heart Cardiac Shape Completion on CT, with an Open Eleven-Structure Statistical Shape Model}

\author{Matej~Gazda\,\orcidlink{0000-0002-8405-7017},
Jakub~Gazda\,\orcidlink{0000-0003-2246-2217},
Juraj~Gazda\,\orcidlink{0000-0002-7334-9540},
and~Peter~Drot\'ar\,\orcidlink{0000-0002-6634-4696}%
\thanks{Matej Gazda is with the Department of Mathematics and Theoretical Informatics, Faculty of Electrical Engineering and Informatics, Technical University of Ko\v{s}ice, Letn\'a 1/9, 042\,00 Ko\v{s}ice, Slovakia (e-mail: matej.gazda@tuke.sk).}%
\thanks{Jakub Gazda is with the 2nd Department of Internal Medicine, Faculty of Medicine, Pavol Jozef \v{S}af\'arik University and L. Pasteur University Hospital, Trieda SNP 1, 040\,11 Ko\v{s}ice, Slovakia.}%
\thanks{Juraj Gazda and Peter Drot\'ar are with the Department of Computers and Informatics, Faculty of Electrical Engineering and Informatics, Technical University of Ko\v{s}ice, Letn\'a 1/9, 042\,00 Ko\v{s}ice, Slovakia.}%
\thanks{This work has been submitted to the IEEE for possible publication. Copyright may be transferred without notice, after which this version may no longer be accessible.}}

\markboth{Gazda \textit{et al.}: Eleven-Structure Cardiac CT SSM and Shape Completion}{Gazda \textit{et al.}: Eleven-Structure Cardiac CT SSM and Shape Completion}

\maketitle

\begin{abstract}
Public cardiac cohorts annotate different subsets of the heart, so shapes from
separate sources cannot be pooled without shared correspondence. Among
released cardiac shape resources, none we identified carries the atrial
appendage, pulmonary veins, and caval stumps as separate blocks in one mesh. Completion benchmarks also compare
deep models against a least-squares projection onto shape modes, not the
conditional estimator the same fitted model implies. We release an
eleven-structure cardiac computed-tomography (CT) statistical shape model,
built from 383 automatically labelled cases in 11\,571-vertex correspondence,
and compare completion estimators under one frozen internal split and endpoint. On a
76-case internal list held out from fitting, a closed-form conditional-Gaussian
estimator reconstructed the missing non-chamber structures at 3.717\,mm mean
per-vertex error, averaged equally over one, three, five, and nine observed
structures. A five-refit mask-conditioned graph variational autoencoder reached
5.248\,mm and nearest-neighbour retrieval 8.931\,mm. The paired difference was
1.531\,mm (95\% confidence interval 1.384 to 1.711), and the
ordering held in a raw-coordinate sensitivity arm. Expert manual labels exist
for 58 external CT cases, but our registered reference is close enough
to score only five structures. There the closed-form estimator again
had lower average surface distance, 95th-percentile Hausdorff distance, and
Chamfer error for both completed atria. On a second public benchmark of 20
cases the reference was close enough for three of four completed structures,
and the same ordering held there. Four structures have no expert reference.
The released model and its completion operator support cohort-unification
research on aligned CT, not clinical use.
\end{abstract}


\IEEEpeerreviewmaketitle

\section{Introduction}

Public cardiac datasets rarely annotate the same parts of the heart. A study that pools two cohorts can therefore be missing a chamber or a vessel that the other cohort provides, and re-annotating the images is seldom an option. This paper asks how accurately the missing structures can be reconstructed from the ones that are present, and which observed structures carry the most information for doing so. Analyses that need the heart as a whole, cardiac simulation~\cite{salvador2024whoreheart, rodero2021linking, strocchi2020publicly, kong2023freeform}, morphometric shape analysis~\cite{bruse2017detecting}, and population-scale phenotyping~\cite{attar2019quantitative}, all work on the cardiac structures as 3D meshes in consistent vertex correspondence, which statistical shape models (SSMs) supply together with population-level variability~\cite{cootes1995active, heimann2009statistical}. Throughout, \emph{whole-heart} means joint coverage of the main cardiac structures rather than of every cardiac tissue: ten of our eleven structures are blood-pool or luminal surfaces and the eleventh is the left-ventricular wall.

Prior work has established ventricular and four-chamber correspondence at scale~\cite{bai2015biventricular, ugurlu2025cardiac, ma2025heartssm}, and has demonstrated four-chamber reconstruction and masked completion; recent additions include sparse-view reconstruction from cine MRI~\cite{liu2026wholeheart}, simulation-oriented template growth~\cite{sveinsson2025meshgrow}, and an age-specific paediatric atlas~\cite{qi2025pediatric}. Section~\ref{sec:related} reviews this literature, and Table~\ref{tab:prior_resources} compares the closest released CT geometry resources. What none of them supplies is the anatomical representation, the baseline design, and the variable-label conditioning that the completion task above requires.

Three gaps follow, and they are what this paper addresses. First, the closest released resources use opening tags or cut rings where this task needs the appendage body, the pulmonary-vein bodies, and the caval stumps as full surface blocks in one shared topology. Second, completion benchmarks compare deep models against a least-squares projection onto shape modes, an estimator weaker than the conditional posterior available from the same fitted model, so the reported margins say more about the choice of baseline than about the deep architecture. Third, existing reconstruction and completion models fix their input and output anatomy to a prescribed acquisition or four-chamber topology, which is the assumption that fails when cohorts are annotated differently from one another.

The fragmentation is concrete. Short-axis cine magnetic resonance imaging (MRI) benchmarks annotate the left ventricle, myocardium, and right ventricle. Other public CT and cardiac magnetic resonance (CMR) benchmarks cover at most seven structures (four chambers, myocardium, aorta, pulmonary artery) and omit the left atrial appendage, pulmonary veins, and venae cavae. TotalSegmentator v2~\cite{wasserthal2023totalsegmentator, wasserthal2025cardiovascular} produces eleven-structure CT segmentations, but most existing cohorts have partial labels. A model that completes the missing structures from whatever subset is available would unify these datasets under one eleven-structure representation.

Cardiac mesh reconstruction methods assume a fixed input and output structure set~\cite{kong2023freeform, ma2025cardiacflow}, and the closest completion work masks chambers during training but emits four-chamber anatomy with no corresponding appendage, pulmonary-vein, or caval blocks~\cite{chen2026vecheart}. The task addressed here is different: one explicit eleven-block topology, conditioned on whichever named blocks a cohort provides.

This paper presents three contributions:
\begin{enumerate}
\item \textbf{A controlled completion benchmark.} Prior cardiac completion work compares against a least-squares projection onto PCA modes. We instead compare standard estimator families under one frozen development/test split, the same named-block panels, and one prespecified geometric endpoint: a regularised conditional-Gaussian posterior, a mask-conditioned graph $\beta$-VAE, a mixture of probabilistic PCAs under prespecified validity rules, a structure-local conditional-Gaussian ablation, and nearest-neighbour and population-mean floors. Two further non-linear arms outside the prespecified family, a published mesh convolution operator substituted into the same architecture and a published completion method run on our own fitted decoders, are reported descriptively as additional non-linear baselines. The individual estimators are established; the contribution is their matched evaluation on eleven-block cardiac correspondence.
\item \textbf{Mask-agnostic completion for cohort unification.} One closed-form operator accepts any coordinate subset for which the regularised conditional solve is defined, without mask-specific retraining. We evaluate named-block panels at $k\in\{1,3,5,9\}$ and axial partial-vertex masks. A fixed biventricular CARE panel supplies a worked cohort-unification example. The contribution is reduced geometric error under the tested aligned-CT masks, not a validated clinical measurement.
\item \textbf{An open eleven-structure correspondence resource.} We release an SSM built from 383 CT cases (11\,571 corresponding vertices) with its template, displacements, and code. The contribution is the open correspondence across these eleven structures together, not a new segmentation or registration method. Two external CT cohorts, CARE2026 and MM-WHS 2017~\cite{zhuang2019mmwhs}, provide expert labels for seven overlapping structures, of which the reference-closeness rule admits five and three respectively for native comparison; the appendage, pulmonary veins, and caval stumps carry only automatic (silver-standard) labels.
\end{enumerate}

The remainder of this paper is organised as follows. Section~\ref{sec:related} reviews related literature. Section~\ref{sec:pipeline} describes the data, mesh extraction, and ANTs SyN registration that produce the eleven-structure SSM. Section~\ref{sec:models} defines the conditional-Gaussian and graph $\beta$-variational-autoencoder ($\beta$-VAE) estimators and their prespecified comparators. Section~\ref{sec:results} reports the evaluation protocol, frozen-split internal completion, external expert-surface and structure-wise evidence, and the geometry, support, truncation, and uncertainty audits. Section~\ref{sec:discussion} discusses interpretation and limitations, and Section~\ref{sec:conclusions} concludes.

\section{Related Work}
\label{sec:related}

\subsection{Cardiac Statistical Shape Models}

Bai et~al.~\cite{bai2015biventricular} built a 1\,000-case biventricular SSM from high-resolution cardiac MRI. Qi et~al.~\cite{qi2025pediatric} recently added an age-specific biventricular atlas from 50 cardiac MRI studies in children aged 10--18 years. Ugurlu et~al.~\cite{ugurlu2025cardiac} scaled to ${\approx}55\,000$ biventricular meshes with open tooling, and Ma et~al.~\cite{ma2025heartssm} extended dynamic modelling to four chambers on ${\approx}96\,000$ UK Biobank participants.

The closest public CT resources contain more anatomy than the shorthand ``four-chamber model'' suggests; Table~\ref{tab:prior_resources} documents them resource by resource. Rodero et~al.~\cite{rodero2021linking} and Strocchi et~al.~\cite{strocchi2020publicly} release aortic and pulmonary-artery walls with their four-chamber models, but the appendage, pulmonary veins, and venae cavae enter as tags or cut rings rather than as separate surface bodies, and Hoogendoorn et~al.~\cite{hoogendoorn2013atlas} built a detailed multiregion spatio-temporal CT atlas from 138 sequences without a reusable public package. The distinction relevant here is a task-enabling choice of representation, not an anatomical count or a qualitative first: every subject uses the same face graph and vertex indices within each block, with eleven named blocks concatenated so their joint covariance can be fitted, whereas these models emit no LAA, PV, SVC, or IVC surface body. None of them can therefore be scored on the eleven-block panel without refitting its own correspondence pipeline, so our positioning against the shape-modelling literature is Table~\ref{tab:prior_resources} on representation, together with the probabilistic-PCA (PPCA) and structure-local conditional-Gaussian family fitted here on identical data.

\begin{table}[t]
\caption{Closest multistructure CT geometry resources. ``Public package''
means a direct public download of a reusable model or mesh cohort, assessed
from the cited publications and their linked repositories at the time of
writing; boundary tags or cut rings are not counted as separate anatomical
surface bodies. None of the prior resources supports missing-label
conditioning; this work conditions on any observed structure-block subset.}
\label{tab:prior_resources}
\centering
\scriptsize
\renewcommand{\arraystretch}{1.03}
\setlength{\tabcolsep}{3pt}
\begin{tabular}{p{0.17\columnwidth}p{0.40\columnwidth}p{0.32\columnwidth}}
\toprule
Resource & Anatomy outside the four chamber walls & Public package \\
\midrule
Hoogendoorn et~al.~\cite{hoogendoorn2013atlas} & Detailed multiregion whole-heart atlas, including great vessels & Atlas and SSM reported; no reusable model package located \\
Rodero et~al.~\cite{rodero2021linking} & AO and PA walls; PV and caval openings tagged; LAA body omitted & SSM parameters and 1\,000 synthetic finite-element meshes \\
Strocchi et~al.~\cite{strocchi2020publicly} & AO and PA walls; LAA, PV, SVC and IVC at cut rings & 24 patient-specific finite-element meshes \\
This work & AO, PA, LAA, PV, SVC and IVC as separate surface blocks in one shared topology & Template, per-case displacement fields and reconstruction code \\
\bottomrule
\end{tabular}
\end{table}

\subsection{Whole-Heart Mesh Reconstruction}

Image-domain restoration and classification are separate stages from geometry extraction: DenoMamba~\cite{ozturk2026denomamba} denoises low-dose CT with fused spatial and channel state-space modules before any geometry is extracted, and shrunken-feature COVID-19 classification from X-ray and CT~\cite{ozturk2021shrunken} labels whole images without vertex correspondence or geometry prediction.

Template-deformation approaches require complete volumetric input~\cite{lin2025pixels}. Kong and Shadden~\cite{kong2023freeform} predict biharmonic control handles from CT segmentations to deform a template into seven structures; their earlier MeshDeformNet~\cite{kong2021meshdeformnet} predicts per-vertex template deformations with graph convolutions. Pak et~al.~\cite{pak2023patient} predict whole-heart meshes from CT. MeshGrow~\cite{sveinsson2025meshgrow} combines a seven-region cardiac template deformation with stepwise aortic tracking to produce simulation-oriented cardiac and vascular meshes, including an explicit aortic-valve interface. Deng et~al.~\cite{deng2026morphinet} refine biventricular meshes from CMR via graph subdivision. On MRI, Beetz et~al.~\cite{beetz2023pccn} reconstruct biventricular point clouds from sparse cine contours without mesh topology, while Liu et~al.~\cite{liu2026wholeheart} map sparse multi-view cine MRI to temporally coherent four-chamber meshes through differentiable contour rendering. Muffoletto et~al.~\cite{muffoletto2026neural} use neural implicit coordinates for biventricular reconstruction with iterative optimisation at inference. These methods solve image-to-geometry reconstruction for a prescribed acquisition, rather than conditioning a correspondence model on whichever anatomical structures a cohort happens to contain.

\subsection{Shape Completion and Generation}

PCN~\cite{yuan2018pcn} and GRNet~\cite{xie2020grnet} complete spatial geometry from partial point clouds without anatomical part labels. CardiacFlow~\cite{ma2025cardiacflow} uses flow matching for four-chamber 3D+t completion and generation. Cardiac Mesh Flow~\cite{ma2026meshflow} extends flow matching to explicit four-chamber meshes in vertex correspondence and volume-conditioned generation. Qiao et~al.~\cite{qiao2025meshheart} generate 3D+t biventricular meshes from demographics. These generative models establish strong non-linear shape priors. Their conditioning variables and output topology are fixed to the structures in their four-chamber task, whereas our evaluation varies which named anatomical blocks are supplied and scores the blocks withheld from that panel.

Anatomical completion is addressed directly by VecHeart~\cite{chen2026vecheart}, which masks chambers during training and reconstructs four-chamber anatomy from a hybrid vector-set latent decoded to an implicit field. It is the closest task-level precedent, but its four-chamber implicit output cannot represent the appendage, pulmonary veins, or venae cavae as corresponding mesh blocks. No compatible eleven-block implementation or pretrained topology is publicly available for a matched retraining baseline; our non-linear comparators instead use the same eleven-block mesh, split, and observation panels as the linear estimator.

Graph convolutional autoencoders~\cite{ranjan2018coma} model face and body meshes, and $\beta$-VAEs~\cite{higgins2017betavae} learn disentangled latent representations. Biffi et~al.~\cite{biffi2018learning} apply cardiac VAEs to disease classification on 3D cardiac segmentations without completion conditioning or generation.

In cranial modelling, two studies combine a linear shape-model prior with a
learned non-linear correction: Pimentel et~al.~\cite{pimentel2020autoimplant}
refine an SSM fit to a defective skull with a generative adversarial network,
and Milojevic et~al.~\cite{mil2024autoskull} predict a correction to PCA
coefficients with a multilayer perceptron (AutoSkull). Neither targets cardiac
anatomy or conditions on an arbitrary observed-structure subset.

\section{Pipeline: From CT to SSM}
\label{sec:pipeline}

\begin{figure*}[t]
\centering
\includegraphics[width=0.75\textwidth]{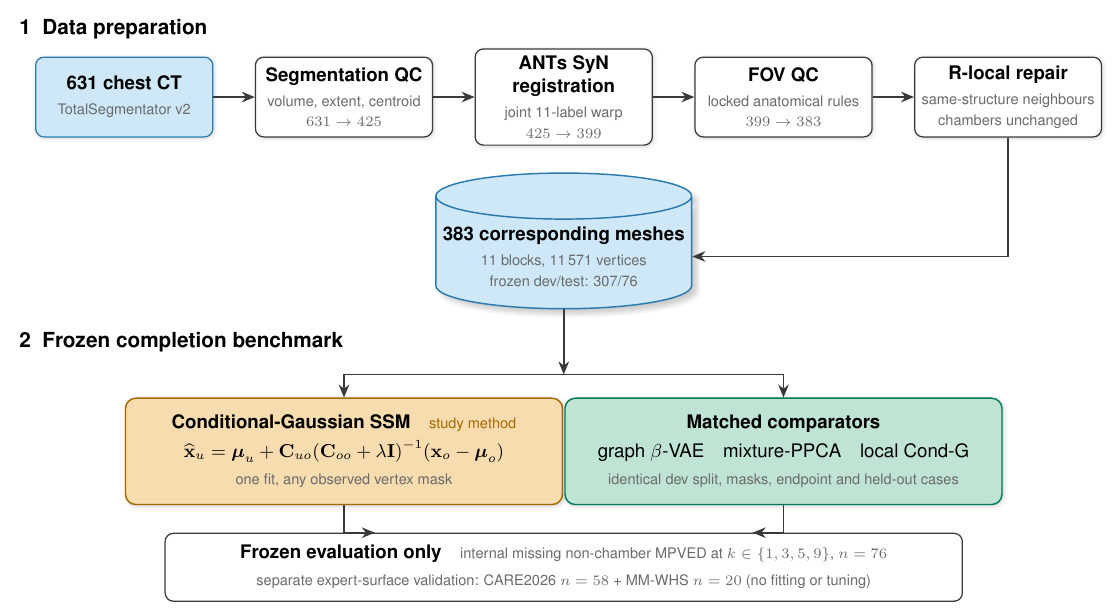}
\caption{End-to-end pipeline and frozen evaluation benchmark. After segmentation QC, joint-label registration, registration/FOV QC, and same-structure R-local repair, the 383 corresponding meshes are divided into a development set used for model selection and a 76-case evaluation set withheld from every fit reported here (the graph architecture predates this partition; Section~\ref{sec:results}). The orange block carries the conditional-mean equation; all comparators use the same split, panels, endpoint, and evaluation cases. CARE2026 and MM-WHS provide separate expert-surface validation without fitting or tuning.}
\label{fig:pipeline}
\end{figure*}

\subsection{Data, Mesh Extraction, and Registration}

This study draws on three public CT cohorts. The first, \emph{internal} in role only (every fit and selection uses it), is 631 volumes from the TotalSegmentator dataset, labelled for the eleven structures of Table~\ref{tab:structures} by TotalSegmentator v2~\cite{wasserthal2023totalsegmentator, wasserthal2025cardiovascular}; it carries silver (automatically generated) labels only and supplies the SSM development and frozen-split internal evaluation cases. The first \emph{external} cohort is the CARE2026 challenge set~\cite{care2026challenge}, sixty volumes across three acquisition subsets (A, B, G), held out entirely from model fitting; seven structures (the four chambers, MYO, AO, and PA) additionally carry expert manual labels, while LAA, PV, SVC, and IVC are silver-only. Two volumes fail an affine check (Table~\ref{tab:cohorts}), leaving $n{=}58$ split 20/18/20 across A, B and G.

The second \emph{external} cohort is all 20 CT training cases of MM-WHS 2017~\cite{zhuang2019mmwhs}, with no exclusions. It provides expert labels for the same seven overlapping structures but none for LAA, PV, SVC, or IVC, and it is held out from every fit and tuning decision, analysed separately as secondary descriptive validation (Section~\ref{sec:eval}). All three cohorts are analysed at 1\,mm isotropic spacing. Before registration, every cohort passes an anatomical orientation check on structure ordering and frame handedness, so that a label map inconsistent with its affine is rejected rather than registered.

Table~\ref{tab:cohorts} summarises the three cohorts: exclusion cascades,
acquisition composition, labels, and split. The internal study mix is
overwhelmingly non-cardiac (336 of 383 cases are thorax--abdomen--pelvis,
trauma, or staging examinations), and electrocardiographic (ECG) gating status and cardiac phase are
absent everywhere; because non-gated acquisitions sample the cardiac cycle
arbitrarily, phase enters as unstructured variation and is treated as a
limitation below. The five fixed development folds determine every
hyperparameter varied in the matched selection grids and the deep-model refit
duration. The graph architecture was fixed before the partition, in an
earlier sweep whose training pool included cases now in the evaluation list
(development-only sensitivity below), and the frozen 76-case list, though
withheld from every fit reported here, is not an independent cohort or strict
confirmatory holdout (Section~\ref{sec:discussion}).

\begin{table}[t]
\renewcommand{\arraystretch}{1.03}
\caption{The three cohorts and the exclusions that produce them. The released
internal metadata supplies age, sex, manufacturer, scanner model, tube
voltage, pathology class, and study type; ``n/a'' attributes are absent from
the released metadata and are not estimated, and ``none'' means no fitting or
tuning. The 76-case evaluation list is frozen (split file released with the artifacts). Both
CARE2026 exclusions fall in subset B; MM-WHS has no exclusions.}
\label{tab:cohorts}
\centering
\scriptsize
\setlength{\tabcolsep}{3pt}
\resizebox{\columnwidth}{!}{%
\begin{tabular}{llll}
\toprule
 & Internal (TotalSegmentator) & CARE2026 & MM-WHS CT \\
\midrule
Role            & SSM fit; internal eval. & indep.\ validation & indep.\ secondary \\
Source cases    & 631 & 60 (A/B/G) & 20 \\
Segmentation QC & $-206$ (vol., extent, centr.) $\rightarrow$ 425 & -- & -- \\
Affine check    & -- & $-2$ (non-orthon.) & -- \\
Registration QC & $-26$ ($>$25\,mm) $\rightarrow$ 399 & -- & -- \\
FOV truncation  & $-16$ & -- & -- \\
\textbf{Retained} & \textbf{383} & \textbf{58} (20/18/20) & \textbf{20} \\
Split/tuning    & 307 dev.\ (5 folds) / 76 withheld$^\dagger$ & none & none \\
Labels          & silver, all 11 & silver 11, expert 7 & expert 7 \\
Scanner models  & 11 distinct & n/a & n/a \\
Median age      & 64 years & n/a & n/a \\
Study mix       & 336 non-cardiac, 47 cardiac & n/a & n/a \\
Contrast phase  & n/a & n/a & n/a \\
ECG gating, phase & n/a, likely mixed & n/a & n/a \\
Spacing analysed & 1\,mm isotropic & 1\,mm isotropic & per case $\rightarrow$ 1\,mm \\
\multicolumn{4}{@{}l}{\scriptsize $^\dagger$The graph architecture predates the partition and may have seen}\\
\multicolumn{4}{@{}l}{\scriptsize cases now in the 76-case evaluation list.}\\
\bottomrule
\end{tabular}}
\end{table}

The end-to-end pipeline is shown in Fig.~\ref{fig:pipeline}. The TotalSegmentator cardiac labels are predominantly luminal: each chamber (LV, RV, LA, RA) is the blood-pool cavity, and the great vessels (AO, PA), pulmonary veins (PV), cavae (SVC, IVC), and left atrial appendage (LAA) are the vessel or appendage lumen. The single exception is the myocardium (MYO), the left-ventricular wall, so the meshes below are blood-pool or luminal surfaces for ten structures and one wall object. From the 631 training volumes, cases were rejected when any core structure (LV, MYO, RV, LA, RA, AO) had non-physiological volume, when the heart bounding box spanned less than 60\,mm craniocaudally, or when structure-centroid separations exceeded 100\,mm, leaving 425 cases.
\begin{table}[t]
	\renewcommand{\arraystretch}{1.03}
	\caption{Eleven cardiac structures and template vertex counts. MYO is the left-ventricular wall; the other ten are blood-pool or luminal surfaces. The model therefore carries no right-ventricular or atrial wall and does not represent total myocardial mass.}
	\label{tab:structures}
	\centering
	\footnotesize
	\begin{tabular}{llrr}
		\toprule
		ID & Structure & Abbrev. & Vertices \\
		\midrule
		1 & Left ventricle & LV & 1\,502 \\
		2 & Myocardium & MYO & 2\,502 \\
		3 & Right ventricle & RV & 2\,002 \\
		4 & Left atrium & LA & 1\,002 \\
		5 & Right atrium & RA & 1\,002 \\
		6 & Aorta & AO & 1\,202 \\
		7 & Pulmonary artery & PA & 802 \\
		8 & Left atrial appendage & LAA & 402 \\
		9 & Pulmonary veins & PV & 602 \\
		10 & Superior vena cava & SVC & 402 \\
		11 & Inferior vena cava & IVC & 151 \\
		\midrule
		& \textbf{Total} & & \textbf{11\,571} \\
		\bottomrule
	\end{tabular}
\end{table}

\begin{figure}[t]
\centering
\includegraphics[viewport=8 46 158 216,clip,width=0.48\columnwidth]{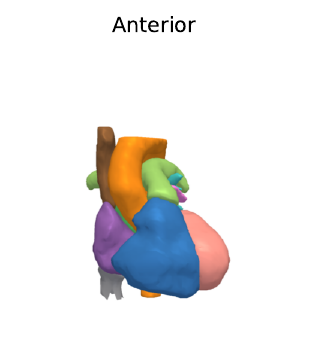}\hfill
\includegraphics[viewport=174 46 324 216,clip,width=0.48\columnwidth]{fig_template_mesh.pdf}\\[2pt]
\includegraphics[viewport=339 46 489 216,clip,width=0.48\columnwidth]{fig_template_mesh.pdf}\hfill
\includegraphics[viewport=505 46 655 216,clip,width=0.48\columnwidth]{fig_template_mesh.pdf}\\[3pt]
\includegraphics[viewport=150 2 311 18,clip,height=17pt]{fig_template_mesh.pdf}\\[1pt]
\includegraphics[viewport=311 2 512 18,clip,height=17pt]{fig_template_mesh.pdf}
\caption{The eleven-structure whole-heart template mesh (11\,571 vertices) in anterior, posterior, left, and right views, each structure a colour-coded surface in fixed vertex correspondence. This single template is warped into every patient's space to build the SSM.}
\label{fig:template}
\end{figure}

Per-structure surface meshes were extracted via marching cubes~\cite{lorensen1987marching} on Gaussian-smoothed label volumes, Laplacian-filtered~\cite{taubin1995signal}, and decimated by quadric edge-collapse to the vertex counts in Table~\ref{tab:structures} (proportional to each structure's mean surface area). The template mesh is extracted from a single selected medoid case (Fig.~\ref{fig:template}); its 11\,571 vertices and fixed connectivity define the correspondence used throughout. Per-case correspondence is obtained purely by warping this template into each patient's space (below), so only the template's own extraction parameters enter the SSM correspondence. The template is the case minimising the sum of Euclidean distances to all others in a 121-dimensional shape descriptor (per-structure centroid, volume, surface area, bounding-box dimensions, principal component analysis (PCA) principal-axis lengths), each feature z-scored across the cohort. Vessel-extent features enter with unit weight; because truncation is common in the cohort, they can pull the medoid toward a truncation-typical case and bake a short-vessel template into the correspondence, one route to the flat distal vessel caps.

Heart centroids vary by up to 870\,mm across scans due to scanner-origin differences. Centre-of-mass (COM) prealignment shifts each case to match the template centroid, reducing median displacement from 275\,mm to 17\,mm. No isotropic-scale or generalised-Procrustes normalisation is applied, so the per-vertex displacement encodes size together with shape rather than shape alone.

Symmetric diffeomorphic registration (ANTs SyN~\cite{avants2008symmetric})
computes a deformation between each patient label image and the template
(cross-correlation similarity, radius~4, iterations 160/80/40). A single joint
warp is estimated on the multi-label volume and propagated to all eleven
structures. The cross-correlation objective is dominated by the large
chambers, so great-vessel correspondence is a by-product of the joint warp
rather than separately optimised; this motivates the representation audit
below. Template vertices are propagated to patient space via a coordinate-map
technique: three scalar images encoding moving-space $x$, $y$, and $z$
coordinates are warped through the SyN transform and sampled trilinearly at
template vertex locations.

Validity-mask propagation prevents undefined coordinate samples: a ones-image
is warped with the same transform, and a template vertex with sampled validity
below 0.5 is left at its centre-of-mass-prealigned template position. The
resulting unsmoothed coordinate-map displacement is retained as the
\emph{raw} sensitivity representation.

The primary \emph{R-local} representation repairs only detached components of
an individual non-chamber surface. Within each structure, an edge is cut when
its patient-space length is both greater than 15\,mm and greater than 3.5 times
its template length. If this disconnects the graph, the largest component is
the anchor and every other component is filled by inverse-distance weighting
from up to 12 nearest anchor vertices in template space (all anchor vertices
are used if fewer than 12 remain). A no-regression guard restores the raw
structure unchanged if a candidate repair increases its longest edge. The five chamber
blocks are copied bit-for-bit: detection thresholds and replacement values use
neither chamber displacement nor any other structure.

Two repair settings enter the evaluation: the primary R-local rule and the
raw arm, which applies no repair and measures sensitivity to omitting it.
R-local changes 868 vertices in 39 cases (mean 0.0196\% of vertices per case;
median zero; maximum 0.467\%); by structure the changed counts (fraction of
all case-vertex opportunities) are AO 259 (0.0563\%), PA 245 (0.0798\%), LAA
98 (0.0637\%), PV 141 (0.0612\%), SVC 104 (0.0675\%), and IVC 21 (0.0363\%),
and chamber counts are exactly zero by construction, so no completion-input
information enters the targets.

The IVC template is trimmed at the $z$-minimum of the descending aorta to remove the variable abdominal segment, reducing IVC vertices from 402 to 151 and the total mesh from 11\,822 to 11\,571. The trim plane is fixed once in template space and not recomputed per case, so the short-stub IVC seen on external CARE2026 is consistent with the fixed template extent rather than evidence of a prediction-specific truncation.

Cases with mean chamber-vertex displacement (over LV, MYO, RV, LA, RA)
exceeding 25\,mm relative to the template are rejected, flagging scans the SyN
cannot recover, typically extreme rotation or partial field-of-view (FOV)
anatomy. This leaves 399 of the 425 segmentation-QC survivors (93.9\%);
dropping a further 16 cases with FOV truncation gives the clean cohort of
$n{=}383$ (90.1\% of the 425 survivors; 383 of the 631 raw volumes). Shapes are
represented as per-vertex displacements from the centre-of-mass-prealigned
template.

Per-structure warp-overlap Dice similarity coefficient (the moving segmentation pushed through the SyN transform and compared to the template label) measures how well the diffeomorphic transform brings each patient into template correspondence, not anatomical accuracy against a manual reference. It was recorded for the CARE2026 external cohort ($n{=}58$) only. Table~\ref{tab:dice} reports the per-structure distribution. The four chamber medians exceed 0.97 and MYO reaches 0.962 (mean of the five per-chamber medians 0.974; per-case pooled chamber mean 0.954); this is not comparable to the segmentation-accuracy Dice of prior whole-heart methods on CT~\cite{kong2021meshdeformnet, zhuang2019mmwhs}, which measure overlap against a reference label. Aorta, SVC, and IVC are systematically lower (medians 0.82, 0.73, 0.53) because external CT FOVs frequently truncate these vessels. This Dice certifies the forward label warp (moving$\to$template), whereas the SSM correspondence uses the inverse (template$\to$moving) field; the two agree closely for a well-behaved diffeomorphism, so the 0.97 figure is an upper proxy for the uncharacterised inverse-consistency residual.

Because the registration target is a TotalSegmentator silver label rather than a manual one, we separately checked agreement against the seven expert-labelled CARE2026 structures (Dice in patient space, no registration). Median overlap is high for the chambers and pulmonary artery, and low for the aorta (0.64), where the expert label covers a shorter aortic extent than the silver label, so the unshared segment is charged as disagreement; the chamber distributions have a noisy low tail (per-case minima to 0.45). This check characterises one upstream layer and neither validates point correspondence nor replaces the reference-closeness rule.

\begin{table}[t]
\renewcommand{\arraystretch}{1.03}
\caption{Per-structure warp-overlap Dice (SyN-warped patient segmentations vs template), CARE2026 ($n{=}58$): a registration-consistency measure, not accuracy against a manual reference. Structures marked $^{\dagger}$ (LAA, PV, SVC, IVC) are silver-only. \textbf{Chambers} is the mean of the five per-chamber medians. p25 is the 25th percentile.}
\label{tab:dice}
\centering
\footnotesize
\begin{tabular}{lcc|lcc}
\toprule
Structure & Median & p25 & Structure & Median & p25 \\
\midrule
LV   & 0.981 & 0.979 & AO   & 0.824 & 0.787 \\
MYO  & 0.962 & 0.944 & PA   & 0.969 & 0.910 \\
RV   & 0.974 & 0.965 & LAA$^{\dagger}$  & 0.941 & 0.938 \\
LA   & 0.978 & 0.977 & PV$^{\dagger}$   & 0.918 & 0.891 \\
RA   & 0.975 & 0.971 & SVC$^{\dagger}$  & 0.725 & 0.619 \\
     &       &       & IVC$^{\dagger}$  & 0.528 & 0.461 \\
\midrule
\multicolumn{3}{l}{\textbf{Chambers}: 0.974} & \multicolumn{3}{l}{\textbf{All 11}: 0.889} \\
\bottomrule
\end{tabular}
\end{table}

The headline external numbers are pooled over the three acquisition subsets. Subset B is the hardest domain (lowest, most variable chamber Dice) and is also where the two excluded non-orthonormal-affine cases fall, so the pooled result is mildly optimistic on the registration-consistency axis, though the silver-label faithfulness it rests on is not specific to subset B.

\section{Completion Models}
\label{sec:models}

The completion problem is the same in every model below. A case is a vector of per-vertex displacements from the template, an observed set of structures fixes some of those vertices, and the model must supply the rest; the models differ only in how the conditional distribution of the unobserved vertices is represented. The conditional-Gaussian posterior accepts any coordinate subset for which the regularised solve is defined, so the same fitted model serves the named-block panels and axial partial-vertex masks evaluated here without retraining; this interface is broader than the tested mask distributions, and we do not infer performance outside them. The models below are standard estimator families used to create a controlled completion benchmark, not claimed as new algorithms.

\begin{figure}[t]
\centering
\includegraphics[width=\columnwidth]{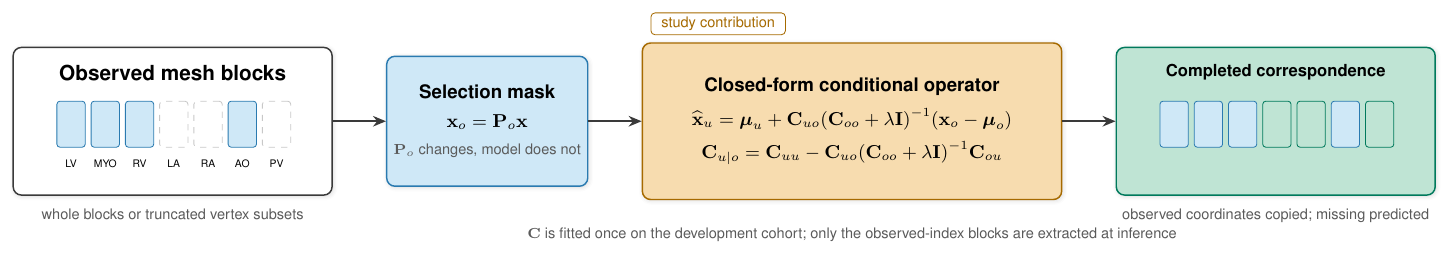}
\caption{The completion operator. A selection matrix $\mathbf{P}_o$ extracts whichever structure blocks or partial vertex sets are observed. The covariance is fitted once; inference extracts the corresponding covariance blocks and evaluates the conditional mean. Observed coordinates are copied to the output; missing coordinates are predicted. The figure writes
the operator in covariance form for readability; the implementation
evaluates the low-rank coefficient-space posterior of Eq.~(1), whose ridge
acts on the posterior precision rather than on $\mathbf{C}_{oo}$.}
\label{fig:completion_operator}
\end{figure}

\subsection{Structure-Conditioned Graph $\beta$-VAE}
\label{sec:completion_vae}

Let $\mathbf{d} \in \mathbb{R}^{V \times 3}$ denote the displacement field of a registered cardiac mesh ($V{=}11\,571$). Given a subset $\mathcal{S} \subset \{1, \ldots, 11\}$ of observed structures, the task is to reconstruct the complete displacement $\hat{\mathbf{d}}$ for all eleven structures. The model extends CoMA (a convolutional mesh autoencoder)~\cite{ranjan2018coma} with a mask-conditioning channel: the encoder takes $[\mathbf{d}_\text{masked}; \mathbf{m}] \in \mathbb{R}^{V \times 4}$, displacements at observed vertices (zero elsewhere) plus a binary mask, and stacks Chebyshev graph convolutions~\cite{defferrard2016chebnet} of order $K{=}3$ with mesh pooling across four resolutions ($11\,571 \to 2\,909 \to 741 \to 210$ vertices), widening from 16 to 128 channels, to a $d_z$-dimensional Gaussian latent. The decoder mirrors the encoder; a differentiable Laplacian output smoothing and a Laplacian-coordinate loss term act only on the deep model.

During training, each sample retains $k \sim \text{Uniform}[k_\text{min}, k_\text{max}]$ complete structure blocks (uniformly one to ten); the rest are zeroed and the mask updated. This covers the named-block panels in the primary evaluation but not arbitrary within-structure surfaces, so the fixed axial partial-vertex masks are a separate out-of-training-distribution sensitivity. The loss combines masked-weighted reconstruction (weight 2 on completed vertices), a Kullback--Leibler (KL) term warmed up linearly to the candidate $\beta_{\max}$, and Laplacian, edge-length, normal-alignment, and inter-structure repulsion regularisers with fixed weights; the released configuration carries the exact values.

The matched search crosses $d_z\in\{16,32,64,128\}$ with
$\beta_{\max}\in\{10^{-4},10^{-3},10^{-2}\}$ on five development folds and
training seeds 42/43, under one fixed Adam recipe with early stopping. The
selected cell is refitted on all 307 development cases for the upper median
of its ten best-checkpoint epoch indices plus one, with no validation-free
checkpoint selection during refit. At evaluation, each refit uses the
deterministic posterior mean, and coordinates from seeds 42--46 are averaged
before the evaluation endpoint of Section~\ref{sec:eval} is computed.

\subsection{Conditional-Gaussian and Classical Shape Baselines}
\label{sec:classical_models}
We complete the shape with the closed-form Bayesian posterior over PCA
coefficients, which, unlike an unregularised least-squares projection,
conditions on the prior over those coefficients. Writing the normalised
flattened displacement as $\mathbf{x} \in \mathbb{R}^{3V}$ with development
mean $\boldsymbol{\mu}$, principal basis $\mathbf{U}$, retained mode variances
$\boldsymbol{\Lambda}$, and residual variance $\sigma^2$, and treating the
coefficients as $\mathbf{z} \sim \mathcal{N}(\mathbf{0},
\boldsymbol{\Lambda})$, the posterior mean given observed coordinates
$\mathbf{x}_o$ is
\begin{equation}
\label{eq:posterior}
\hat{\mathbf{z}} = \left( \tfrac{1}{\sigma^2}\mathbf{U}_o^{\top}\mathbf{U}_o + \boldsymbol{\Lambda}^{-1} \right)^{-1} \tfrac{1}{\sigma^2}\mathbf{U}_o^{\top}(\mathbf{x}_o - \boldsymbol{\mu}_o),
\end{equation}
and the completed shape is $\boldsymbol{\mu} + \mathbf{U}\hat{\mathbf{z}}$,
with observed coordinates copied to the final output. The implementation adds
a relative diagonal ridge
$r\,\overline{\operatorname{diag}(\mathbf{A})}\mathbf{I}$ to the posterior
precision $\mathbf{A}$ in \eqref{eq:posterior}. Residual variance is the mean
of the discarded eigenvalues within the centred training rank $N-1$, bounded
below by $10^{-10}$ for numerical safety; it is a regulariser, not a calibrated
measurement-noise estimate. We call this estimator conditional Gaussian
(Cond-G).

Global Cond-G searches $M\in\{50,100,150,200,240\}$ and
$r\in\{10^{-4},10^{-3},10^{-2},10^{-1}\}$ within each representation arm.
Within 0.01\,mm of the lowest development completion error, the rule chooses fewer modes
and then the larger ridge. Both raw and R-local select $M=200$ and
$r=10^{-4}$.

Mixture PPCA~\cite{tipping1999mixtures} is fitted in the lossless
$N-1$-dimensional fold-PCA coefficient space. A diagonal-covariance Gaussian mixture model (GMM) on the
first 20 coefficients uses $K\in\{1,2,3,5\}$, seeds 0--2, and PPCA ranks
$q\in\{20,40,80,160\}$. A cell is invalid if the GMM fails to converge, an
effective component contains fewer than 30 cases, or, for $K>1$,
$q>\lfloor n_{\mathrm{eff,min}}/3\rfloor$. Invalid cells are reported and not
rescued. The local Cond-G challenger selects the nearest
$k_{\mathrm{NN}}\in\{25,50,100,200,\mathrm{all}\}$ training shapes by RMS
distance over observed coordinates, fits a local posterior with
$q\in\{20,50,100,200\}$, and rejects $q>k_{\mathrm{NN}}-1$. Both local searches
select the unrestricted all-neighbour, $q=200$ limit; therefore no distinct
local model is promoted to the held-out endpoint.

\section{Results}
\label{sec:results}

\subsection{Experimental Setup}
\label{sec:eval}

The primary internal endpoint uses the frozen 76-case list only after every
development choice has been locked. We evaluate $k\in\{1,3,5,9\}$ observed
structures with 20 fixed panels per $k$ (panel seed 42). Five $k=9$ panels
observe every non-chamber structure and contain no defined primary target; the
same five are omitted for every method, leaving panel counts 20/20/20/15. For
a case and panel, the endpoint is mean per-vertex Euclidean distance (MPVED)
over all missing non-chamber vertices; panels are averaged within case and
$k$, and the four $k$-specific case values receive equal weight in the primary
aggregate. An equal-structure endpoint prevents larger meshes from dominating
a secondary descriptive analysis.

Every compatible estimator is selected on the same five development folds,
panels, normalisation, and four-$k$ MPVED criterion, over the search grids of
Section~\ref{sec:models}. Mixture PPCA and local Cond-G use prespecified
validity rules rather than repairing invalid cells after inspection. Full
grids, tie rules, and failure accounting are given with the released
artifacts. The all-development mean shape and observed-coordinate
nearest-neighbour retrieval are descriptive floors, not selected models.

R-local is the primary representation and the raw coordinate-map field is a
prespecified sensitivity arm.
Hyperparameters are selected separately within raw and R-local development
folds. The primary predeclared contrast is R-local $\beta$-VAE minus R-local
Cond-G, equally averaged over the four $k$ values; positive values favour
Cond-G. Because the graph architecture predates the evaluation partition, its paired
bootstrap interval and resampling-floor $p$ value are descriptive conditional
on this shared-pool design rather than strict confirmatory inference.
Exactly seven supportive contrasts share one Holm family: the analogous raw
four-$k$ contrast, the four R-local per-$k$ contrasts, R-local PPCA minus
Cond-G at the selected $K=1$, $q=160$ limit, and the R-local
local-minus-global structural identity. The last is retained under the family
map frozen before development selection resolved and is zero by construction,
because selection chose the unrestricted global limit and the scorer aliases
the same rank-correct prediction; retaining it is conservative for the
remaining Holm-adjusted contrasts. All other comparisons are descriptive or
exploratory.

Inference resamples cases, not vertices or panels. A single frozen
$10{,}000\times76$ index matrix, released with the artifacts, carries every paired method,
panel, structure, representation, and training seed together. We report
two-sided percentile 95\% intervals and paired-bootstrap achieved-significance
levels; the sole primary contrast is unadjusted and the seven supportive
$p$ values are Holm-adjusted. A prespecified escalation
would add refit seeds 47--51 in both representation arms if the absolute
primary difference fell below the five-refit seed SD.

We audit uncertainty exploratively in the R-local domain on four fixed nested
observation panels: LV alone; LV/MYO/RV; all five chambers; and those five
blocks plus AO/PA/LAA/PV. Cond-G uses its exact conditional covariance under the selected
finite-rank Gaussian model, residual variance, and ridge; for a fixed panel
this covariance is homoscedastic across cases, not a learned case-adaptive
confidence estimate. For the $\beta$-VAE, five refits each provide 32
posterior draws; the primary matrix is the empirical second moment of all 160
decoded draws about the deterministic five-refit ensemble used for efficacy,
with a separately labelled sensitivity recentred on the Monte Carlo mean.
Coverage of 50\%, 80\%, and 95\% three-dimensional $\chi^2$ ellipsoids,
Gaussian negative log likelihood, and the ratio of case-mean squared error to
case-mean covariance trace are reported over each panel's missing non-chamber
vertices and over the common SVC/IVC intersection, with descriptive
case-bootstrap intervals and no tests. A $10^{-6}$\,mm$^2$ eigenvalue floor is
used only for inverses and log determinants, and its activation is reported.

A separate descriptive support audit fits the selected Cond-G PCA subspace on
the 307 complete development shapes and assigns each test case its full-shape
Mahalanobis distance in that subspace. The tail is fixed once as distance
strictly above the R-local development 95th percentile, reusing the same case
set for the raw arm; typical/tail distributions, tail-minus-typical
case-bootstrap intervals, rank association, and five prespecified worst cases
are reported without $p$-values or exclusion.

CARE2026 native prediction surfaces are compared with expert labels by
area-weighted symmetric average surface distance (ASSD), the 95th-percentile Hausdorff distance (HD95), and the Chamfer root-mean-square (RMS) distance,
beside the registered-reference diagnostic and a prespecified reference-closeness rule: a structure receives native
metrics only where its registered reference lies within fixed distance
limits of the expert surface. MM-WHS supplies a separate 20-case expert-labelled CT
validation with no retuning, exhaustive seven-label panels at $k=1,3,5$, and
its own case bootstrap, released with the artifacts; its intervals and contrasts are
descriptive and never pooled with CARE2026 or the internal endpoint.
Geometry, CARE expert-surface, and fixed axial AO/IVC truncation experiments
use two fixed predictions, the deterministic Cond-G fit and the coordinate
mean of the five deterministic $\beta$-VAE refits, in raw and R-local
coordinates.

\subsection{Internal Completion on the Frozen Evaluation Split}
\label{sec:fresh_completion}

Development selection chose Cond-G with $M=200$ modes and relative ridge $10^{-4}$
in both representations. The raw $\beta$-VAE winner was
$(d_z,\beta_{\max})=(64,10^{-2})$, refit for 233 epochs; the R-local winner
was $(128,10^{-2})$, refit for 226 epochs. The five deterministic refit
predictions are averaged as coordinate fields before the evaluation endpoint of Section~\ref{sec:eval} is computed. No
$K>1$ mixture met the prespecified validity rules of Section~\ref{sec:classical_models}; the only
eligible winner is the single-component PPCA limit at $q=160$. The local
Cond-G search selected all development cases and $q=200$, exactly the global
prediction, so it is retained as an identity check rather than a distinct
estimator.

Table~\ref{tab:crossover} gives all frozen-split means. In the primary
R-local representation, four-$k$ MPVED is 3.717\,mm for Cond-G against
5.248\,mm for the $\beta$-VAE, and Cond-G has the lowest value among the
non-identical compatible estimators at every $k$; the separately selected raw
arm gives the same ordering.

\begin{table*}[t]
\renewcommand{\arraystretch}{1.03}
\caption{Frozen 76-case internal evaluation completion MPVED (mm).
Each $k$ entry is the case mean over its fixed panels; the four-$k$ column
equally weights the four $k$ values, with a case-bootstrap 95\% CI. PPCA is
the selected $K{=}1,q{=}160$ limit; local Cond-G reuses the exact global
prediction by construction.}
\label{tab:crossover}
\centering
\scriptsize
\setlength{\tabcolsep}{3pt}
\begin{tabular}{llccccc}
\toprule
Representation & Method & $k=1$ & $k=3$ & $k=5$ & $k=9$ & Four-$k$ [95\% CI] \\
\midrule
\multirow{6}{*}{Raw}
& Mean shape & 11.218 & 11.278 & 11.189 & 10.909 & 11.149 [10.310, 12.040] \\
& Nearest neighbour & 10.096 & 9.031 & 8.508 & 8.644 & 9.070 [8.621, 9.563] \\
& Cond-G & \textbf{5.479} & \textbf{3.918} & \textbf{3.193} & \textbf{3.003} & \textbf{3.898} [3.629, 4.206] \\
& $\beta$-VAE & 6.596 & 5.238 & 4.693 & 4.748 & 5.319 [4.991, 5.701] \\
& PPCA ($K=1$) & 5.524 & 3.958 & 3.248 & 3.064 & 3.948 [3.683, 4.251] \\
& Local Cond-G$^\dagger$ & 5.479 & 3.918 & 3.193 & 3.003 & 3.898 [3.629, 4.206] \\
\midrule
\multirow{6}{*}{R-local}
& Mean shape & 11.157 & 11.215 & 11.130 & 10.858 & 11.090 [10.245, 11.990] \\
& Nearest neighbour & 9.942 & 8.904 & 8.352 & 8.527 & 8.931 [8.503, 9.405] \\
& Cond-G & \textbf{5.304} & \textbf{3.732} & \textbf{3.007} & \textbf{2.825} & \textbf{3.717} [3.480, 3.984] \\
& $\beta$-VAE & 6.522 & 5.172 & 4.621 & 4.677 & 5.248 [4.935, 5.613] \\
& PPCA ($K=1$) & 5.348 & 3.767 & 3.061 & 2.882 & 3.764 [3.530, 4.027] \\
& Local Cond-G$^\dagger$ & 5.304 & 3.732 & 3.007 & 2.825 & 3.717 [3.480, 3.984] \\
\bottomrule
\end{tabular}
\vspace{2pt}

{\scriptsize $^\dagger$Structural identity, not an independently fitted held-out result.}
\end{table*}

\begin{table*}[t]
\renewcommand{\arraystretch}{1.03}
\caption{Frozen paired family. Differences are challenger minus Cond-G, so
positive values favour Cond-G. The R-local four-$k$ primary is unadjusted; the
next seven rows are the complete Holm family ($m=7$). The final row is an
identity check rather than empirical evidence.}
\label{tab:significance}
\centering
\scriptsize
\setlength{\tabcolsep}{3pt}
\begin{tabular}{lccc}
\toprule
Contrast & Difference (mm) & 95\% CI (mm) & Multiplicity result \\
\midrule
\textbf{R-local $\beta$-VAE minus Cond-G, four-$k$ primary}
& \textbf{1.531} & \textbf{[1.384, 1.711]} & Unadjusted $p<0.001$ (floor) \\
\midrule
Raw $\beta$-VAE minus Cond-G, four-$k$ & 1.421 & [1.264, 1.610] & $p_{\rm Holm}<0.002$ \\
R-local $\beta$-VAE minus Cond-G, $k=1$ & 1.218 & [1.023, 1.439] & $p_{\rm Holm}<0.002$ \\
R-local $\beta$-VAE minus Cond-G, $k=3$ & 1.440 & [1.298, 1.602] & $p_{\rm Holm}<0.002$ \\
R-local $\beta$-VAE minus Cond-G, $k=5$ & 1.613 & [1.466, 1.800] & $p_{\rm Holm}<0.002$ \\
R-local $\beta$-VAE minus Cond-G, $k=9$ & 1.852 & [1.689, 2.047] & $p_{\rm Holm}<0.002$ \\
R-local PPCA ($K=1$) minus Cond-G, four-$k$ & 0.047 & [0.038, 0.056] & $p_{\rm Holm}<0.002$ \\
R-local local minus global Cond-G, four-$k$ & 0.000 & [0.000, 0.000] & Structural identity \\
\bottomrule
\end{tabular}
\end{table*}

The primary paired $\beta$-VAE-minus-Cond-G difference is $+1.531$\,mm
(95\% CI 1.384--1.711; $p<0.001$ at the paired-bootstrap floor). Every
empirical supportive contrast remains positive after the prespecified Holm
adjustment (Table~\ref{tab:significance}). Individual R-local $\beta$-VAE
refits give differences of 1.791--1.918\,mm (seed SD 0.053\,mm). Averaging
their coordinate fields before the nonlinear Euclidean endpoint reduces the
ensemble difference by 0.314\,mm relative to the mean individual-refit
difference; the two summaries measure different quantities. The frozen seed-escalation rule does
not trigger.

The graph architecture itself is a fixed input to this comparison: its
Chebyshev order, channel ladder, and depth come from an earlier 11-cell sweep
on a previous partition and representation, and because cases later assigned
to the evaluation list could have contributed to that sweep, it is disclosed
only as architecture-design evidence. Before scoring the frozen evaluation
list, we therefore froze a development-only R-local sensitivity at the locked
$d_z=128,\beta_{\max}=10^{-2}$ cell: ten configurations vary Chebyshev order,
width, one additional encoder/decoder block, learning rate, mask weight, and
Laplacian weight over the same five folds and seeds 42/43, entering no
hypothesis family and never scoring an evaluation case; the two seeds match
the development-selection budget of every candidate, while seeds 44--46
belong only to the all-development refit ensemble. The base and eight alternative
configurations with complete finite outputs span 5.983--6.433\,mm development
MPVED, with paired alternative-minus-base means from $-0.011$ to
$+0.439$\,mm; the $K=10$, width-16 configuration is invalid after two
non-finite runs (122 predictions), with no value imputed and no aggregate
computed. This bounded sensitivity neither establishes architectural
robustness nor changes the primary refits.

A further development-only experiment addressed the operator axis. With the
convolution replaced by EdgeConv~\cite{wang2019dgcnn} inside the same architecture, results proved
strongly dependent on the optimisation settings: under the settings shared
with the graph $\beta$-VAE the substituted operator was far behind, while
under operator-specific retuning it reached 5.774\,mm, below the
5.983--6.433\,mm envelope of the $\beta$-VAE configurations tested here and
still above the conditional-Gaussian model's 4.055\,mm on the same folds.
These post-primary runs never score an evaluation case. What they establish is
that the shared training settings are not neutral across architectures, recorded as a limitation of the
matched-protocol design; our conclusions concern the models evaluated here,
not non-linear completion methods in general.

\subsection{Additional Non-Linear Baselines}
\label{sec:secondary_comparators}

To test whether the deep architecture is representative rather than a weak
instance of its class, two descriptive post-lock arms were evaluated on the same
frozen cases, panels, and endpoint (Tables~\ref{tab:secondary}
and~\ref{tab:secondary_care}): SpiralNet++~\cite{gong2019spiralnet}, a published mesh convolution
operator substituted into our architecture at parameter parity (14.155\,M
against 14.2\,M), and latent optimisation following Litany \emph{et al.}~\cite{litany2018deformable}, a
published completion method run on our own locked $\beta$-VAE decoders, so
that pair shares weights and differs only in inference procedure. Latent
optimisation outperforms our deep model by 0.204\,mm [0.169, 0.243]
internally and 0.368\,mm [0.338, 0.403] on MM-WHS, yet trails Cond-G by
1.327\,mm [1.201, 1.482] and 1.426\,mm [1.272, 1.579]; on CARE2026 expert
surfaces the ordering holds a third time, by 0.032\,mm (LA) and 0.091\,mm
(RA) over the $\beta$-VAE, with AO and PA suppressed by the reference-closeness
rule as for the primary methods. That a published completion method beats the
$\beta$-VAE on both cohorts while losing to the closed-form estimator is the
evidence that the deep arm is not a strawman of our construction: latent optimisation fixes the
observed coordinates and asks what the fitted model implies for the
remainder, structurally what the conditional mean does, and it still falls
short.

\begin{table}[t]
\renewcommand{\arraystretch}{1.03}
\caption{Additional non-linear baselines, R-local representation, MPVED (mm) with
case-bootstrap 95\% intervals. The internal column is the frozen 76-case split
at four observed-structure counts; the MM-WHS column is exhaustive panels at
$k=1,3,5$ over the seven expert-overlap structures, so only the ordering
within a column is comparable. Both arms lie outside the prespecified Holm
family and are descriptive; the latent-optimisation setting is not retuned
externally. Cond-G, the $\beta$-VAE and the floors repeat
Table~\ref{tab:crossover} for reference; the same arms on CARE2026 expert
surfaces, a different endpoint, are in Table~\ref{tab:secondary_care}.}
\label{tab:secondary}
\centering
\scriptsize
\setlength{\tabcolsep}{3pt}
\resizebox{\columnwidth}{!}{%
\begin{tabular}{llcc}
\toprule
Method & Type & Internal 76 & MM-WHS ($n=20$) \\
\midrule
Cond-G & Closed-form linear & \textbf{3.717} [3.480, 3.984] & \textbf{4.823} [4.446, 5.226] \\
Latent optimisation$^\ddagger$ & Published method & 5.044 [4.746, 5.389] & 6.248 [5.755, 6.777] \\
$\beta$-VAE & Feed-forward deep & 5.248 [4.935, 5.613] & 6.616 [6.109, 7.152] \\
SpiralNet++$^\S$ & Published operator & 6.710 [6.283, 7.218] & 8.916 [8.105, 9.741] \\
\midrule
Nearest neighbour & Retrieval floor & 8.931 [8.503, 9.405] & 10.907 [10.145, 11.684] \\
Mean shape & Population floor & 11.090 [10.245, 11.990] & 13.174 [12.045, 14.419] \\
\bottomrule
\end{tabular}}
\vspace{2pt}

{\scriptsize $^\ddagger$Litany \emph{et al.}, run on our frozen $\beta$-VAE
decoders; the autoencoder is ours, the completion algorithm is theirs.
$^\S$Operator substituted into our architecture at parameter parity, not a
reproduction of a published end-to-end system.}
\end{table}

\begin{table}[t]
\renewcommand{\arraystretch}{1.03}
\caption{Additional non-linear baselines on CARE2026 expert surfaces, R-local, biventricular
panel, mean ASSD / HD95 / Chamfer RMS (mm) over 58 cases. The endpoint is
distance to expert manual segmentations and is not comparable to
Table~\ref{tab:secondary}. AO and PA are suppressed for every method by the
prespecified reference-closeness rule. Cond-G and the $\beta$-VAE repeat
Table~\ref{tab:care_subsets} for reference.}
\label{tab:secondary_care}
\centering
\scriptsize
\setlength{\tabcolsep}{3pt}
\resizebox{\columnwidth}{!}{%
\begin{tabular}{llcc}
\toprule
Method & Type & LA & RA \\
\midrule
Cond-G & Closed-form linear & \textbf{3.275 / 7.846 / 4.138} & \textbf{3.479 / 9.007 / 4.586} \\
Latent optimisation & Published method & 3.902 / 8.941 / 4.822 & 4.243 / 10.265 / 5.413 \\
$\beta$-VAE & Feed-forward deep & 3.934 / 8.990 / 4.860 & 4.333 / 10.372 / 5.495 \\
SpiralNet++ & Published operator & 4.299 / 9.701 / 5.266 & 4.605 / 10.868 / 5.779 \\
\bottomrule
\end{tabular}}
\end{table}

\begin{figure*}[t]
\centering
\includegraphics[width=0.72\textwidth]{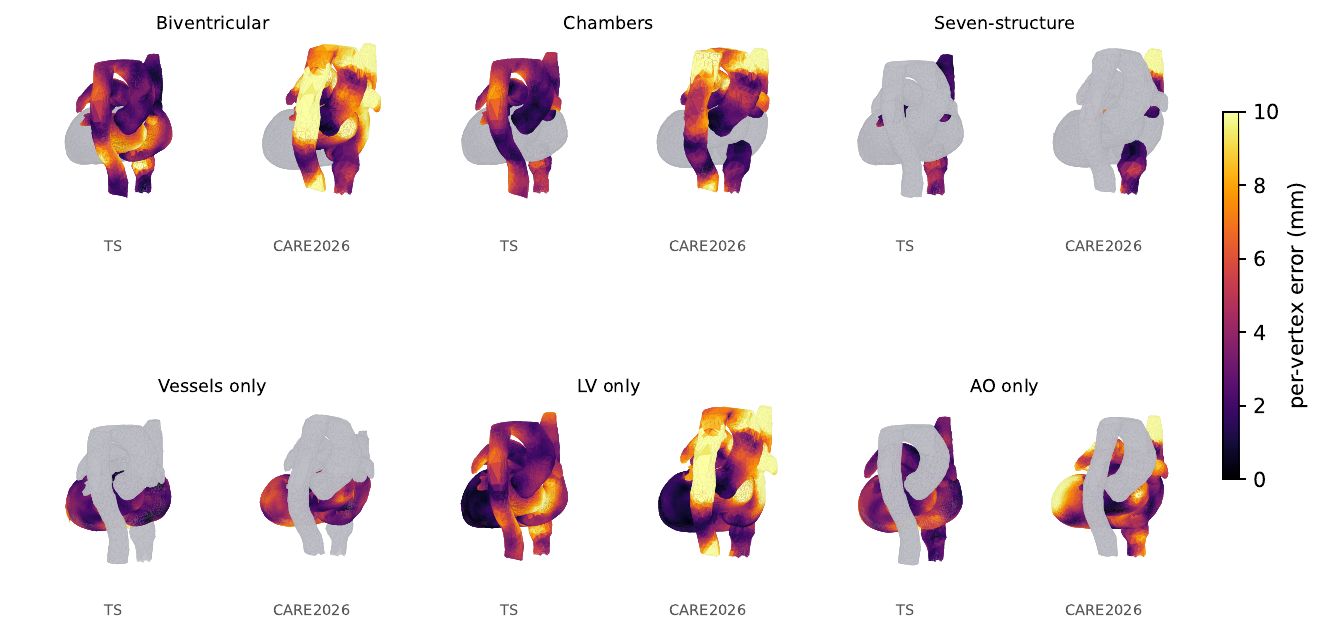}
\caption{Completion across six observed-structure panels, one internal
TotalSegmentator case (TS, s0365) beside one external CARE2026 case
(A\_Case1020), each the median case of its cohort by the displayed Cond-G estimator's mean
missing-vertex error over the six panels. Observed structures are grey; the conditional-Gaussian
model completes the rest, coloured by per-vertex error (mm). The estimator is
the selected configuration of Table~\ref{tab:crossover}, fitted on the 307
development cases in the R-local domain.}
\label{fig:qualitative}
\end{figure*}

Fig.~\ref{fig:qualitative} shows these aggregates on the median case of each
cohort under the six displayed panels, so neither is a best case. The
completed anatomy closes into a coherent whole-heart mesh at every panel,
observed blocks are reproduced exactly because their coordinates are copied,
and the residual concentrates on the vessel and appendage surfaces furthest
from the observed chambers. The single-structure panels (LV only, AO only)
carry visibly the largest residual, the geometric statement behind the
$k$-dependence in Table~\ref{tab:crossover}; the external case shows the same
spatial pattern with a larger residual, consistent with
Section~\ref{sec:external_evidence}.

\subsection{External Validation}
\label{sec:external_evidence}

\textbf{CARE2026.} A completion is scored against an expert surface only
where the registered reference itself lies close to that surface (the
reference-closeness rule of Section~\ref{sec:eval}); otherwise the distance
would measure annotation mismatch rather than completion quality. The rule
passes LV, MYO, RV,
LA, and RA in both representations. Under R-local, their median/p90 reference
ASSD ranges from 0.830/2.201 to 1.510/2.091\,mm. AO fails under its
prespecified longitudinal-coverage mismatch (18.895/21.694\,mm), while PA
misses the strict p90 threshold by 0.026\,mm
(3.398/6.954\,mm versus 6.928\,mm). Thus neither AO nor PA receives a native
accuracy summary, and the successful $1.2\times$ PA threshold sensitivity does
not change that fixed decision. The designated 50\% AO region is also
technically non-evaluable in 46/58 R-local cases.

Tables~\ref{tab:care_subsets} and~\ref{tab:care_loo_subsets} report the
eligible R-local native metrics, pooled and by acquisition subset. With
biventricular input, Cond-G has lower pooled descriptive ASSD, HD95, and
Chamfer RMS means than the $\beta$-VAE for both LA and RA. The leave-one-out
panel shows the same pooled ordering for the five eligible structures, but the
complete A/B/G display is not uniformly ordered in every metric--subset cell:
subset-A RV HD95 is 8.073\,mm for the $\beta$-VAE and 8.349\,mm for Cond-G.
The subsets are descriptive; no subgroup robustness claim or hypothesis test
is made.

\begin{table}[t]
\renewcommand{\arraystretch}{1.06}
\caption{R-local CARE biventricular native prediction-versus-expert surface
metrics, pooled ($n=58$) and by acquisition subset (A/B/G contain 20/18/20
cases). The panel observes LV/MYO/RV; values are mean ASSD / HD95 / Chamfer
RMS in mm. AO and PA are omitted after failing the reference-closeness rule.}
\label{tab:care_subsets}
\centering
\scriptsize
\setlength{\tabcolsep}{3pt}
\begin{tabular}{llcc}
\toprule
Population & Method & LA & RA \\
\midrule
Pooled & Cond-G & 3.275 / 7.846 / 4.138 & 3.479 / 9.007 / 4.586 \\
Pooled & $\beta$-VAE & 3.934 / 8.990 / 4.860 & 4.333 / 10.372 / 5.495 \\
\midrule
A & Cond-G & 2.634 / 6.643 / 3.353 & 3.035 / 8.085 / 3.989 \\
A & $\beta$-VAE & 3.103 / 7.553 / 3.911 & 3.780 / 9.373 / 4.811 \\
B & Cond-G & 4.010 / 9.642 / 5.032 & 4.028 / 10.598 / 5.351 \\
B & $\beta$-VAE & 4.878 / 11.412 / 6.034 & 4.706 / 11.572 / 6.011 \\
G & Cond-G & 3.253 / 7.432 / 4.119 & 3.427 / 8.496 / 4.494 \\
G & $\beta$-VAE & 3.915 / 8.247 / 4.752 & 4.549 / 10.290 / 5.715 \\
\bottomrule
\end{tabular}
\end{table}

\begin{table}[t]
\renewcommand{\arraystretch}{1.00}
\caption{R-local CARE leave-one-expert-structure-out completion, pooled
($n=58$) and by acquisition subset (A/B/G contain 20/18/20 cases). Values are
mean ASSD / HD95 / Chamfer RMS in mm. AO and PA are omitted because their references are too far.}
\label{tab:care_loo_subsets}
\centering
\scriptsize
\setlength{\tabcolsep}{3pt}
\begin{tabular}{llcc}
\toprule
Missing structure & Population & Cond-G & $\beta$-VAE \\
\midrule
LV & Pooled & 2.362 / 6.407 / 3.119 & 3.899 / 9.623 / 4.942 \\
   & A      & 1.818 / 4.744 / 2.358 & 3.367 / 8.177 / 4.254 \\
   & B      & 3.697 / 10.361 / 4.969 & 4.841 / 12.575 / 6.258 \\
   & G      & 1.704 / 4.510 / 2.215 & 3.582 / 8.411 / 4.446 \\
\midrule
MYO & Pooled & 2.056 / 5.667 / 2.729 & 2.943 / 7.583 / 3.769 \\
    & A      & 1.695 / 4.541 / 2.212 & 2.526 / 6.257 / 3.171 \\
    & B      & 2.947 / 8.627 / 4.036 & 3.847 / 10.332 / 5.046 \\
    & G      & 1.616 / 4.129 / 2.068 & 2.547 / 6.434 / 3.219 \\
\midrule
RV & Pooled & 2.717 / 8.098 / 3.770 & 3.377 / 8.555 / 4.335 \\
   & A      & 2.868 / 8.349 / 3.969 & 3.016 / 8.073 / 3.994 \\
   & B      & 3.247 / 10.314 / 4.619 & 4.148 / 10.495 / 5.305 \\
   & G      & 2.088 / 5.852 / 2.807 & 3.045 / 7.290 / 3.801 \\
\midrule
LA & Pooled & 2.417 / 6.228 / 3.192 & 3.393 / 7.837 / 4.207 \\
   & A      & 2.169 / 5.919 / 2.871 & 2.577 / 6.451 / 3.266 \\
   & B      & 3.023 / 7.642 / 3.919 & 4.445 / 10.407 / 5.495 \\
   & G      & 2.120 / 5.264 / 2.860 & 3.262 / 6.909 / 3.989 \\
\midrule
RA & Pooled & 2.800 / 7.434 / 3.763 & 3.878 / 9.530 / 4.988 \\
   & A      & 2.577 / 7.013 / 3.458 & 3.474 / 8.729 / 4.448 \\
   & B      & 3.246 / 9.006 / 4.441 & 4.372 / 10.762 / 5.607 \\
   & G      & 2.621 / 6.439 / 3.456 & 3.836 / 9.224 / 4.972 \\
\bottomrule
\end{tabular}
\end{table}

\textbf{MM-WHS.}\label{sec:mmwhs_transfer} MM-WHS supplies a second independent CT cohort (20 cases), with no exclusions,
pooling, or retuning. As on CARE2026, the reference-closeness rule decides which
structures receive native summaries: it admits three of the four completed
structures.
Under R-local, median/p90 reference ASSD is 0.628/0.687\,mm for LA,
0.620/0.838\,mm for RA, and 2.884/4.724\,mm for PA, all within the fixed
3.464/6.928-mm limits. AO fails at 17.060/19.024\,mm, and the directional
decomposition identifies annotation scope rather than registration error as
the cause: where MM-WHS labels the aorta, the registered reference agrees
with the expert surface to 1.075\,mm averaged over the 20 cases, comparable to
LA and RA, while the opposite direction reaches 32.558\,mm because the
reference extends beyond the annotated ascending segment into the arch and
descending aorta, and a symmetric surface distance charges that unannotated
extent as error. The same mismatch fails AO on CARE2026 (18.895\,mm), so AO
receives no native accuracy summary in either external cohort. The rule is
symmetric by design and was not modified for this structure; the favourable
one-directional value is a diagnostic, not partial validation. PA passes only
in the R-local representation; in the raw arm it reaches 9.999\,mm and fails,
so its eligibility is representation-dependent and reported as such.

For the three eligible structures, Cond-G has lower mean ASSD than the
$\beta$-VAE on LA (2.854 versus 3.989\,mm), RA (2.733 versus 3.435\,mm), and
PA (4.895 versus 5.441\,mm), and the same ordering holds for HD95
(7.570/7.647/15.041 versus 9.742/9.147/16.579\,mm) and Chamfer RMS
(3.704/3.655/6.963 versus 4.995/4.467/7.652\,mm). Single-component PPCA is not
separable from Cond-G at this cohort size (2.896/2.753/4.928\,mm); both remain
well below the nearest-neighbour (6.474/5.263/8.092\,mm) and population-mean
(6.673/6.628/8.064\,mm) floors. The structure-local Cond-G ablation is
numerically identical to Cond-G here, with a paired difference of zero,
because the observed panel is exactly LV, MYO, and RV; it separates only on
the internal eleven-block panels. These are descriptive means over 20 cases
from a segmentation challenge's training split: they corroborate the internal
ordering on an independent cohort without supporting a confirmatory claim.

MM-WHS also transfers the internal representation-space endpoint. Against the
seven-label registered correspondence target, R-local Cond-G gives 4.823\,mm
MPVED (95\% descriptive case-bootstrap interval 4.446--5.226), the
$\beta$-VAE 6.616\,mm (6.109--7.152), and single-component PPCA 4.863\,mm
(4.487--5.265), averaged equally over $k=1,3,5$; the nearest-neighbour and
population-mean floors give 10.907 and 13.174\,mm, the corresponding raw
values are 5.502, 6.963, and 5.512\,mm, and the paired $\beta$-VAE minus
Cond-G contrast is 1.794\,mm (1.623--1.961). These are representation-space
transfer results, not native expert-surface accuracy.

\textbf{Evidence by structure.}\label{sec:structure_evidence}

Table~\ref{tab:structure_status}, fixed before frozen-split selected-model
scoring, answers three questions for every structure separately: whether its
reference is an expert manual label or an automatic silver label, whether
that reference is close enough to score against, and what evidence exists.
The internal endpoint scores only missing non-chamber blocks, so LV, MYO,
RV, LA, and RA receive no internal completion score. For each of the six non-chamber structures (AO, PA, LAA, PV, SVC, IVC) at
each observed-structure count $k\in\{1,3,5,9\}$, 24 combinations in all,
Cond-G has a lower descriptive mean error than the $\beta$-VAE, and the same
holds in all 24 raw-representation combinations. The four $k$ values are kept
separate in the released artifacts, and no cross-$k$ average, reliability
tier, or composite rank is formed, so no summary can hide a weak structure.

\begin{table}[t]
\renewcommand{\arraystretch}{1.15}
\caption{The evidence behind each structure; structures with identical
status share a row. ``Reference close enough'' means the registered reference lies
within the fixed distance limits of the expert surface, so a native
comparison is meaningful; the internal
benchmark scores only structures that are completion targets there.
``Cond-G better'' is the lower descriptive mean error against the
$\beta$-VAE at every $k$. AO's reference is too far because the expert
annotations cover a shorter aortic extent than the whole-aorta reference,
not because of misregistration (Section~\ref{sec:external_evidence}).}
\label{tab:structure_status}
\centering
\scriptsize
\setlength{\tabcolsep}{3pt}
\begin{tabular}{p{0.16\columnwidth}p{0.22\columnwidth}p{0.24\columnwidth}p{0.24\columnwidth}}
\toprule
Structures & Internal benchmark & CARE2026 expert check & MM-WHS expert check \\
\midrule
LV, MYO, RV & never a completion target & reference close enough (leave-one-out panel) & expert label, but observed as input \\
LA, RA & never a completion target & reference close enough (both panels) & reference close enough; Cond-G better \\
AO & Cond-G better at every $k$ & reference too far & reference too far \\
PA & Cond-G better at every $k$ & reference too far & reference close enough under R-local; Cond-G better \\
LAA, PV, SVC, IVC & Cond-G better at every $k$ & no expert label & no expert label \\
\bottomrule
\end{tabular}
\end{table}

CARE thus provides expert-surface evidence with a close-enough reference for five structures, MM-WHS for
three of its four completed targets, and four structures have no external
expert reference at all; AO fails the reference-closeness rule in both external cohorts, PA is
eligible on MM-WHS only under R-local, and the subset-A RV HD95 reversal
prevents a uniform per-metric claim. Fixed-topology and FOV caveats are
strongest for AO extent, PA distal branching, variable PV branching, and the
capped SVC/IVC stumps. These absences and qualifications are results, not
values to be averaged away.

\subsection{Repair Ablation and Validity Audits}
\label{sec:audits}

R-local modifies only 868 of 4,431,693 internal case--vertex opportunities,
in 39/383 cases, and no chamber coordinate. Omitting it raises the four-$k$
Cond-G mean from 3.717 to 3.898\,mm and the $\beta$-VAE mean from 5.248 to
5.319\,mm; Cond-G remains favoured in both arms, which preserves the selected
model ordering without establishing anatomical truth. The same fixed
transform intervenes more on both external cohorts: on CARE2026 it modifies
2,174 of 671,118 opportunities in all 58 cases (per-case mean 0.324\%, median
0.251\%, maximum 1.193\%), and on MM-WHS 640 of 231,420 opportunities in all
20 cases (per-case mean 0.277\%, median 0.216\%, maximum 1.219\%), versus a
0.0196\% internal per-case mean; we treat that disparity as
preprocessing-domain sensitivity, not successful external correction.

Both selected models degrade outside development support. Among the 12 cases
above the frozen R-local development 95th-percentile full-shape Mahalanobis
threshold, Cond-G increases from 3.335 to 5.758\,mm
(tail penalty 2.423\,mm; 95\% descriptive interval 1.693--3.193) and the
$\beta$-VAE from 4.774 to 7.778\,mm (3.004\,mm; 1.967--4.329).
Continuous distance/error Spearman correlations are 0.883 and 0.790,
respectively. No direct interval or test was prespecified for the difference
between penalties, so no differential robustness claim is made. The distance
uses the complete target, is unavailable at inference, and is not a marker of
pathology.

True triangle tests contradict a collision-free interpretation. Across 76
R-local biventricular completions, Cond-G/$\beta$-VAE/target cases with
self-intersection include LA 6/17/10, RA 4/31/14, AO 8/11/17, and PA 7/9/6.
Any true triangle contact with observed LV, MYO, or RV occurs in 34/63/3 cases
for AO and 45/56/6 for PA; frequent LA--LV and RA--RV contacts also occur in
the targets because these structures have shared anatomical interfaces.
Predicted degenerate-face counts are zero, but triangle contacts are not
overlap volume and zero degeneracy does not imply anatomical validity.

Finally, fixed axial AO/IVC retention at 0, 25, 50, 75, and 100\% evaluates
only completion after template registration; hidden-set composition changes
with the cut and several raw and R-local error curves are non-monotone, so the
experiment establishes neither monotonic truncation robustness nor
modality-specific partial-surface registration performance.

\subsection{Downstream Measurement, Robustness, and Uncertainty}
\label{sec:downstream}

Three further analyses, each with its plan fixed before the numbers were
seen, address what a user would take from a completed mesh, whether the fitted
variation looks like cardiac phase, and what happens when an observed
structure is only partly present. The plans and result files are released with
the reproducibility artifacts.

\textbf{Morphometric retention.} From the biventricular panel we complete the
remaining structures and compute per-structure enclosed volume for the seven structures with usable
volumes (the trimmed IVC stub is excluded), comparing the completion against
the case's own registered mesh
(Table~\ref{tab:morphometry}). Because a completion is a deterministic
function of the observed structures, the question is how much of a measurement
survives, so each structure is also compared against three alternatives
available without a shape model: the development-set median volume, the mean
shape, and a development-only ridge regression of the missing log volume on
the three observed chamber volumes.
Improvement is claimed only when the paired case-bootstrap interval for the
difference in absolute error excludes zero. Internally, five of seven structures
beat all three alternatives; pulmonary-vein volume is not separated from any of
them ($-0.01$\,mL, $[-0.31, +0.28]$ against the regression) and caval volume is
not separated from the regression. Externally the atria retain their advantage
over the regression while AO, PA, and PV are not separated from any baseline, and
caval volume is worse than a constant ($+2.49$\,mL, $[+1.65, +3.35]$). Atrial
sphericity improves on the mean shape internally for both atria ($-0.021$ and
$-0.028$) but degrades externally for LA ($+0.011$, $[+0.004, +0.018]$).
Multiplicative limits of agreement run $\times[0.75, 1.36]$ for LA and widen to
$\times[0.67, 1.59]$ for SVC, and calibration slopes fall to 0.64 (PV) and 0.73
(SVC), that is, reversion towards the mean for the smallest structures. Volume
is one scalar summary of a surface and no clinical tolerance is defined here,
so these are measurement-agreement results, not clinical validation.

\textbf{Anatomy and phase.} Phase and anatomy are not identifiable from one
static scan per subject, so we do not decompose them; two model-side
observations bound the concern. Within the first 20 principal-component coefficients (a fixed analysis
subspace, distinct from the $M{=}200$ completion model),
log ventricular and log atrial volume correlate \emph{positively} ($r=0.549$
[0.439, 0.630] under R-local, $0.528$ [0.425, 0.610] raw; 400 model-refitting
bootstrap resamples), the signature of overall chamber size rather than the
reciprocal covariance expected from phasic opposition, and the leading mode
tracks patient age (Spearman 0.493). Completion error varies with a
left-ventricular cavity-fraction proxy ($-13.8$\,mm per unit [$-26.1$,
$-2.3$]); the proxy conflates phase with dilation, hypertrophy, and
segmentation quality, so the association is reported without a cause.

\textbf{Partial observation of an input structure.} The fixed axial AO/IVC
retention curves above truncate structures that are themselves completion
targets, so we added a like-for-like experiment in which one \emph{observed}
chamber is axially truncated at 100, 75, 50, 25, and 0\% of its extent while
the hidden set is held fixed, from the inferior end as primary and the
superior end as sensitivity. Non-inferiority was declared in advance at a
1.531\,mm margin, the paired margin by which Cond-G beats the graph
$\beta$-VAE. Against a fully observed panel at 4.599\,mm, truncating LV
changes nothing measurable even when it is removed entirely ($-0.004$\,mm,
one-sided upper bound 0.018\,mm), MYO removal costs 0.037\,mm, and RV, the
only arm with a visible trend, costs 0.295\,mm (upper bound 0.392\,mm). All
six primary cells (three truncated chambers, on the internal and CARE
cohorts) are non-inferior at the margin, in both cut directions and in the
raw arm; the regularised condition number rises only
from $2.5\times10^{2}$ to $4.3\times10^{2}$ across the masks and a ridge
sensitivity at $10^{-5}$ and $10^{-3}$ changes nothing, so this is
insensitivity rather than the regulariser dominating. Under the tested masks and margin, the left-ventricular blood pool
contributes nothing measurable once its wall is observed. The truncation is applied to meshes already in correspondence,
built from complete anatomy, so this is completion from partial input rather
than registration of a cropped image.

\begin{table}[t]
\renewcommand{\arraystretch}{1.03}
\caption{Morphometric retention from the biventricular panel: per-structure
volume agreement between the completion and the case's own registered mesh.
MAE is mean absolute error; medAPE is median absolute percentage error;
$\Delta$ is the paired difference in absolute error against the
development-only regression on the three observed chamber volumes, with its
95\% case-bootstrap interval (negative favours completion). Vessel and
appendage rows are segment volumes under the fixed template crop.}
\label{tab:morphometry}
\centering
\scriptsize
\setlength{\tabcolsep}{3pt}
\begin{tabular}{lcccc}
\toprule
& \multicolumn{3}{c}{Internal ($n{=}76$)} & CARE2026 ($n{=}58$) \\
\cmidrule(lr){2-4}\cmidrule(lr){5-5}
Structure & MAE (mL) & medAPE & $\Delta$ [95\% CI] & MAE (mL), medAPE \\
\midrule
LA  & 7.99  & 8.3\%  & $-6.84$ [$-9.43$, $-4.43$] & 18.47, 18.5\% \\
RA  & 9.61  & 8.9\%  & $-7.29$ [$-10.36$, $-4.35$] & 20.89, 25.7\% \\
AO  & 13.81 & 5.6\%  & $-7.99$ [$-12.75$, $-3.33$] & 23.08, 13.1\% \\
PA  & 6.74  & 8.5\%  & $-4.51$ [$-6.35$, $-2.68$] & 13.39, 17.5\% \\
LAA & 0.93  & 10.7\% & $-0.49$ [$-0.71$, $-0.27$] & 1.41, 14.8\% \\
PV  & 1.41  & 13.1\% & $-0.01$ [$-0.31$, $+0.28$] & 1.55, 13.7\% \\
SVC & 3.18  & 12.3\% & $-0.60$ [$-1.30$, $+0.21$] & 5.52, 22.6\% \\
\bottomrule
\end{tabular}
\end{table}

\textbf{Model and uncertainty boundaries.}\label{sec:model_boundaries}

No $K>1$ PPCA mixture met the prespecified validity rules, and the
local search selects its all-neighbour global limit; neither supplies evidence
for a distinct nonlinear or local advantage. The bounded graph sensitivity
cannot exclude other operators, pooling hierarchies, losses, or larger
cohorts, and the axial partial-vertex masks above are the only
within-structure masks evaluated.

Neither uncertainty construction is calibrated on the 76 evaluation cases.
Cond-G is overconservative: across the four observation panels, its nominal
50\% and 95\% ellipsoids actually cover 0.927--0.988 and 0.991--0.999 of the
missing non-chamber vertices, with error-to-covariance-trace ratios of
0.097--0.293. The $\beta$-VAE is overconfident: its 160-draw empirical second
moment covers only 0.026--0.052 and 0.124--0.203, with ratios 4.448--10.054.
The direction persists on the fixed SVC/IVC population, the alternate Monte
Carlo centre is nearly identical, and no covariance eigenvalue reaches the
numerical floor. These are cohort-level diagnostics under two non-equivalent
constructions; they provide neither a causal explanation of miscoverage nor
per-patient or clinical confidence.

The two constructions nevertheless fail in opposite directions, and the
direction matters more than the magnitude for any reliability claim. Crossing
the audit with the frozen out-of-support strata of Section~\ref{sec:audits}
separates the 76 evaluation cases into 64 typical and 12 atypical; both
artifacts were fixed before any result was seen. Table~\ref{tab:uncertainty_strata}
reports nominal 95\% coverage by stratum. Cond-G covers 1.000 on typical
anatomy in every panel and still 0.944 to 0.999 on the atypical stratum, while
its nominal 50\% coverage falls from 0.978--0.996 down to 0.651--0.949 and its
error-to-trace ratio rises from 0.072--0.199 up to 0.229--0.863, toward one
rather than past it: its conservatism is consumed, not idle, on the cases
furthest from the development distribution. The $\beta$-VAE covers 0.128 to
0.205 on typical and 0.105 to 0.194 on atypical cases, with the panel-1 ratio
rising from 9.177 to 12.163, so it is overconfident everywhere and most
overconfident where the anatomy is unusual. Root-mean-square error roughly
doubles across the strata for both models, from 5.609 to 11.105\,mm for Cond-G
and 6.206 to 10.638\,mm for the $\beta$-VAE on panel 1, and the same
directions hold on the anatomy-fixed SVC/IVC population. This exploratory,
post-hoc stratification rests on 12 atypical cases and does not make either
construction calibrated. What it supports is narrower and still useful: the
target-derived Mahalanobis distance acts as a retrospective reliability flag
for where the reported regions are least trustworthy, and the linear estimator fails in the
conservative direction while the deep ensemble fails in the overconfident one.

\begin{table}[!t]
\caption{Nominal 95\% ellipsoid coverage by out-of-support stratum, R-local, 76
evaluation cases (64 typical, 12 atypical). Strata come from the frozen
development 95th-percentile Mahalanobis threshold. Exploratory and post-hoc;
brackets are descriptive case-bootstrap intervals.}
\label{tab:uncertainty_strata}
\centering
\footnotesize
\setlength{\tabcolsep}{4pt}
\resizebox{\columnwidth}{!}{%
\begin{tabular}{lcccc}
\toprule
 & \multicolumn{2}{c}{Cond-G} & \multicolumn{2}{c}{$\beta$-VAE} \\
\cmidrule(lr){2-3}\cmidrule(lr){4-5}
Observed panel & typical & atypical & typical & atypical \\
\midrule
LV & 1.000 & 0.946 [0.868, 0.998] & 0.128 & 0.105 [0.071, 0.138] \\
LV, MYO, RV & 1.000 & 0.944 [0.878, 0.991] & 0.129 & 0.140 [0.104, 0.174] \\
Five chambers & 1.000 & 0.974 [0.953, 0.993] & 0.141 & 0.139 [0.103, 0.175] \\
Nine structures & 1.000 & 0.999 [0.996, 1.000] & 0.205 & 0.194 [0.110, 0.281] \\
\bottomrule
\end{tabular}}
\end{table}

\section{Discussion}
\label{sec:discussion}

The main finding is deliberately narrow: in this registered CT displacement
space, with matched development selection and a frozen 76-case evaluation
split drawn from the same source pool as development (not an independent cohort),
Cond-G improves the four-$k$ non-chamber MPVED over the selected graph
$\beta$-VAE by 1.531\,mm. The raw-coordinate sensitivity, each individual $k$,
five refits, and the eligible CARE native summaries support the same overall
ordering but answer different questions and are not pooled. The much smaller
0.047-mm internal difference from the eligible single-component PPCA limit
shows that much of the advantage is shared by regularised single-component
Gaussian conditioning rather than unique to one implementation. The result
positions Cond-G against matched alternatives rather than the wider
literature (Section~\ref{sec:eval}).

One plausible explanation is that centre-of-mass alignment, joint SyN
registration, and a fixed template absorb enough pose and correspondence
variation that a regularised Gaussian conditional mean is a strong estimator
at this cohort size; this is a hypothesis about the constructed
representation, not evidence that cardiac anatomy is Gaussian. The invalid
multi-component cells, the global limit selected by the local search, and the
bounded architecture sensitivity cannot distinguish insufficient sample size
from model misspecification, nor exclude a better nonlinear model,
registration strategy, or loss.

The external cohorts sharpen rather than erase the boundary. CARE provides
eligible native expert-surface evidence for five structures and shows lower
pooled Cond-G metrics in the registered biventricular and leave-one-out
settings, but one subset--metric cell reverses and AO/PA fail the reference-closeness rule.
MM-WHS preserves the descriptive registered-target ordering without retuning
and admits LA, RA, and PA natively, where Cond-G is again lower on all three
metrics; AO fails there as on CARE, and PA is eligible only under R-local.
MM-WHS corroborates the internal ordering while showing that measurability is
decided per structure and representation rather than per dataset; at 20 cases
it is corroboration, not confirmatory validation.

The raw/R-local comparison carries its own caution: R-local is sparse
internally and its omission does not reverse model ordering, but the higher
intervention prevalence on MM-WHS reveals domain sensitivity. Separating
silver-target MPVED, warp-overlap diagnostics, registered-reference
measurability, and native expert distances prevents one favourable layer from
certifying another.

The concrete downstream use demonstrated here is cohort unification: the fixed
biventricular CARE panel completes structures that an external cohort never
labelled, putting cases with different label inventories into one eleven-block
correspondence space where per-structure morphometry can be computed on a
common vertex set. The example is deliberate and bounded.

Finally, coordinate accuracy is not geometric or clinical validity.
Self-intersections and cross-structure triangle contacts remain, ostial
distortion and volumetric overlap between completed and observed structures
are not quantified, measurement agreement degrades for the smallest blocks
(pulmonary-vein volume is not separated from a constant, and external caval
volume is worse than one), uncertainty ellipsoids are miscalibrated in
opposite directions, and the axial
truncation curves do not establish robustness. The released meshes and
completion operator are research resources for controlled
correspondence-space experiments; this study does not validate individual
anatomical measurements, abnormality detection, clinical decisions,
haemodynamic or electrophysiological simulation, or completion from
unregistered clinical images.

\subsection{Limitations}

First, the internal 307/76 partition is a frozen evaluation split drawn from
the same source pool used to develop the representation and model families,
not an independent cohort. The reported fits are isolated from those 76
outcomes, but the graph architecture predates the partition and was selected
in a sweep whose training pool included cases now in the evaluation set, so
the paired bootstrap summaries are descriptive conditional on this design
rather than strict confirmatory inference; only the development-only bounded
sensitivity is clean of that issue. CARE and MM-WHS are independent but small
($n=58$ and $n=20$) and their analyses are descriptive. We are not aware of a
larger public CT cohort carrying expert labels for all eleven structures, so
the external benchmarks answer the independence half of validation, not the
scale half.

Second, the internal endpoint is error to a TotalSegmentator-derived,
SyN-registered silver target, not error to anatomical truth. CARE supplies
expert surfaces for seven structures but the reference-closeness rule of
Section~\ref{sec:external_evidence} admits five, MM-WHS admits three (one
only under R-local), and LAA, PV, SVC, and IVC have no external expert
surface (Table~\ref{tab:structure_status}). The fixed-topology and FOV
caveats of Section~\ref{sec:structure_evidence} apply, warp-overlap Dice is
insensitive to tangential vertex sliding, and single-template propagation may
introduce template bias.

Third, the cohort is CT-only, a convenience sample with unknown health,
scanner, ancestry, and geographic composition, and ungated: a single static
scan cannot separate within-cycle motion from inter-subject variation, so the
released mean and modes are not a normative healthy reference, same-instant
input and target do not remove phase mixture from the population
distribution, and cross-phase conditioning (such as predicting an
end-systolic atrium from an end-diastolic ventricle) is outside the evaluated
distribution and untested. The proxy analyses of Section~\ref{sec:downstream} (concordant
chamber covariance, an age-linked leading mode) are consistent with
size-driven variation but measure geometry, not timing, and cannot exclude a
phase contribution; the cavity-fraction association is reported without an
identified cause.

Fourth, the model comparison is finite and contains no published external
system, because none emits the eleven blocks the endpoint scores
(Section~\ref{sec:related}). The graph search matches latent width
and KL weight, the development-only sensitivity varies nearby architecture
choices and one alternative operator, and that experiment showed a shared
training protocol is not neutral across architectures: the comparison is
controlled rather than architecture-optimal, and a matched protocol is no
guarantee of fairness. The pooling hierarchy, alternative objectives,
extensive per-architecture optimisation, and substantially larger datasets
remain untested, and invalid $K>1$ mixture cells are missing evidence, not
evidence that multimodal models fail. Health status and diagnosis are
unavailable in all three cohorts, so no pathology stratum can be formed; the
target-derived Mahalanobis tail is the feasible substitute, measures
degradation away from development support rather than on disease, and cannot
serve at inference.

Fifth, the outputs remain point completions. Neither tested uncertainty
construction is calibrated and no conformal or external calibration layer is
evaluated: a layer fitted on the 307 development cases would be invalid
because those cases already determined the fits and the selected
configurations, and CARE2026 and MM-WHS are too small at 58 and 20 cases to
spare an independent calibration partition. The stratified coverage of
Section~\ref{sec:model_boundaries} locates where the stated regions fail but
does not repair them: it is marginal over cases rather than
patient-conditional, rests on 12 atypical cases, quantifies only each
construction's geometric spread about its own prediction, and excludes
silver-label, registration, and expert-annotation error and cohort shift.
Nonzero self-intersections and contacts, variable topology, open surfaces,
and non-monotone truncation behaviour preclude collision-free, watertight,
simulation-ready, or clinical-use claims.

\section{Conclusions}
\label{sec:conclusions}

In the frozen 76-case internal evaluation split, matched Cond-G
completion outperforms the selected graph $\beta$-VAE by 1.531\,mm across four observed-
structure counts; the raw sensitivity and eligible CARE native results retain
the overall descriptive ordering. Evidence is nevertheless incomplete by
structure: CARE admits five native targets, MM-WHS three of four, AO fails the
reference-closeness rule in both external cohorts, and four structures lack external expert
surfaces. Sparse
representation repair, support, geometry, truncation, and uncertainty audits
expose additional domain and validity limits. We therefore release the
383-case eleven-structure CT correspondence resource and the Cond-G baseline
as research tools for correspondence-space experiments on aligned CT, not
for clinical use.

\section*{Reproducibility Statement}

The public repository at \url{https://github.com/BraveDistribution/openheart-ssm} provides the eleven-structure template mesh, the core CT-to-correspondence pipeline, and model code under the MIT licence. A versioned reproducibility archive accompanying this article provides the exact analysis scripts, frozen split and bootstrap files, manifests, and machine-readable result tables. The per-case R-local displacement fields and reconstructed meshes will be deposited in a versioned public archive at publication. All source image data are public and de-identified; no new data were collected and no image data are redistributed. The reported results were produced with TotalSegmentator 2.13.0 (silver labels), ANTsPy 0.6.3 (registration), and a CUDA 12.8 build of PyTorch (graph models).

\section*{Acknowledgements}

This work is funded by the EU NextGenerationEU through the Recovery and Resilience Plan for Slovakia under the project No.\ 09I03-03-V04-00394. The authors acknowledge the Technical University of Ko\v{s}ice for providing high-performance computing (HPC) resources on the Perun cluster.

\bibliographystyle{IEEEtran}
\bibliography{bibliography}

\raggedbottom

\begin{IEEEbiography}[{\includegraphics[width=1in,height=1.25in,clip,keepaspectratio]{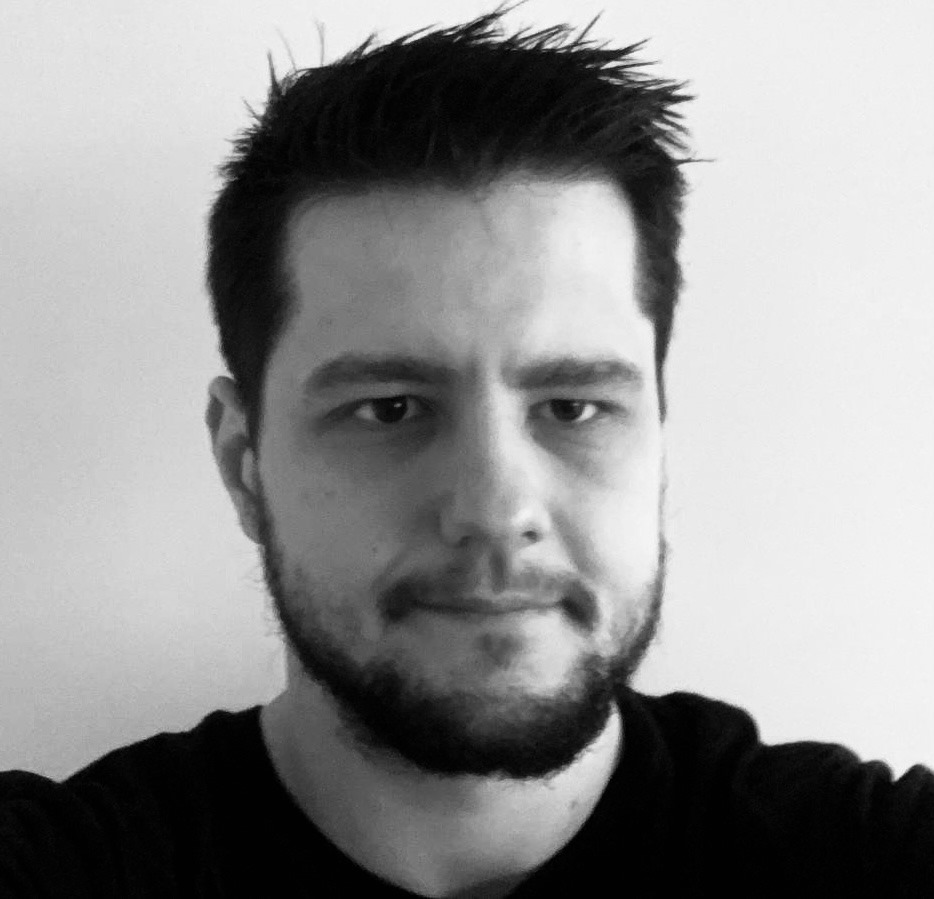}}]{Matej Gazda}
received the B.Sc.\ and M.Sc.\ degrees in computer modelling from the Technical University of Ko\v{s}ice, Slovakia, in 2017 and 2019, respectively, and the Ph.D.\ degree in informatics from the same university in 2023. He is currently with the Department of Mathematics and Theoretical Informatics, Faculty of Electrical Engineering and Informatics, Technical University of Ko\v{s}ice. His research interests include the application of artificial intelligence to medical imaging.
\end{IEEEbiography}

\begin{IEEEbiography}[{\includegraphics[width=1in,height=1.25in,clip,keepaspectratio]{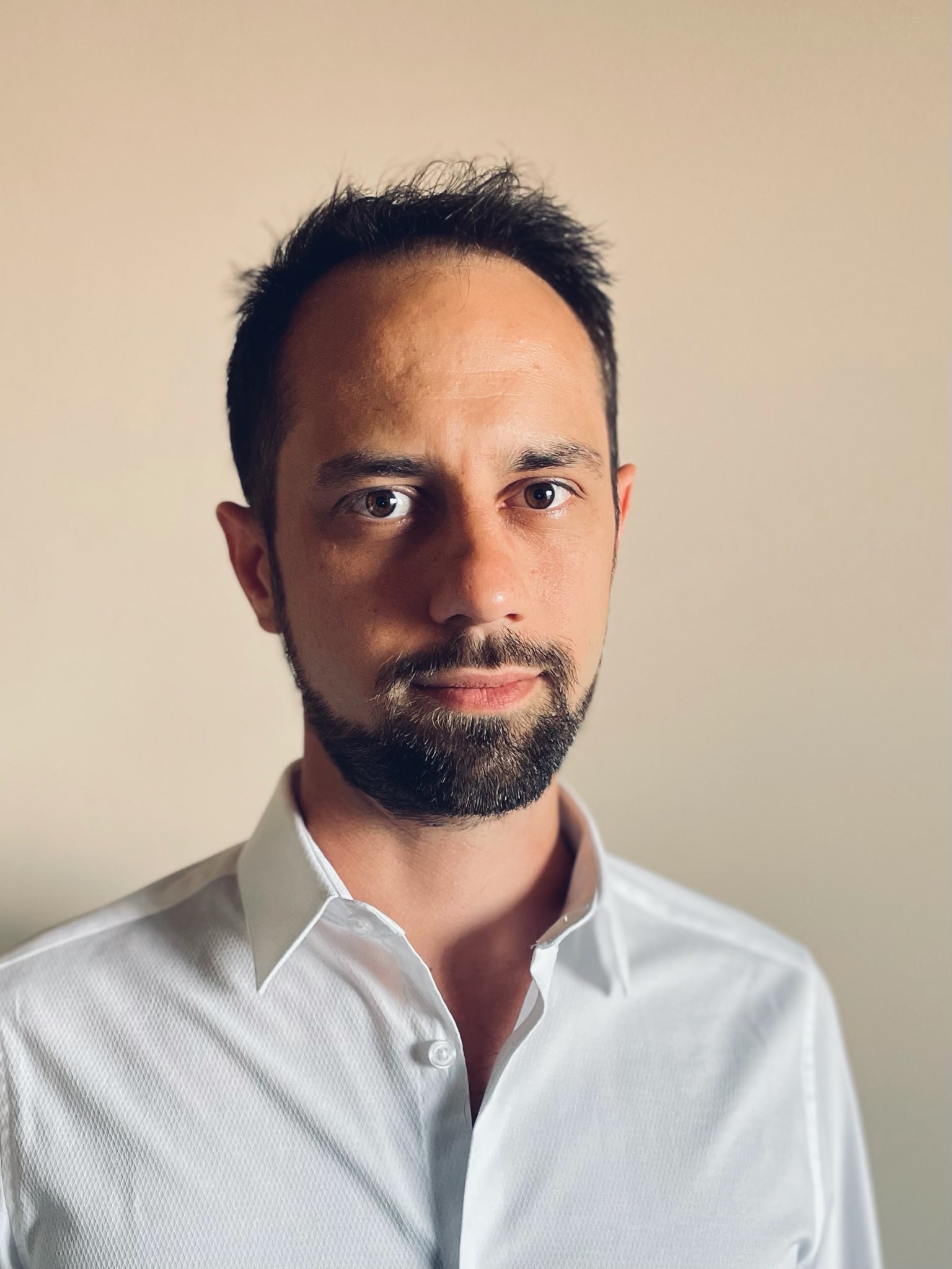}}]{Jakub Gazda}
is a physician and researcher with the Faculty of Medicine, Pavol Jozef \v{S}af\'arik University and L.\ Pasteur University Hospital, Ko\v{s}ice, Slovakia. His research interests are in hepatology and gastroenterology, including the epidemiology and prognosis of primary biliary cholangitis.
\end{IEEEbiography}

\begin{IEEEbiography}[{\includegraphics[width=1in,height=1.25in,clip,keepaspectratio]{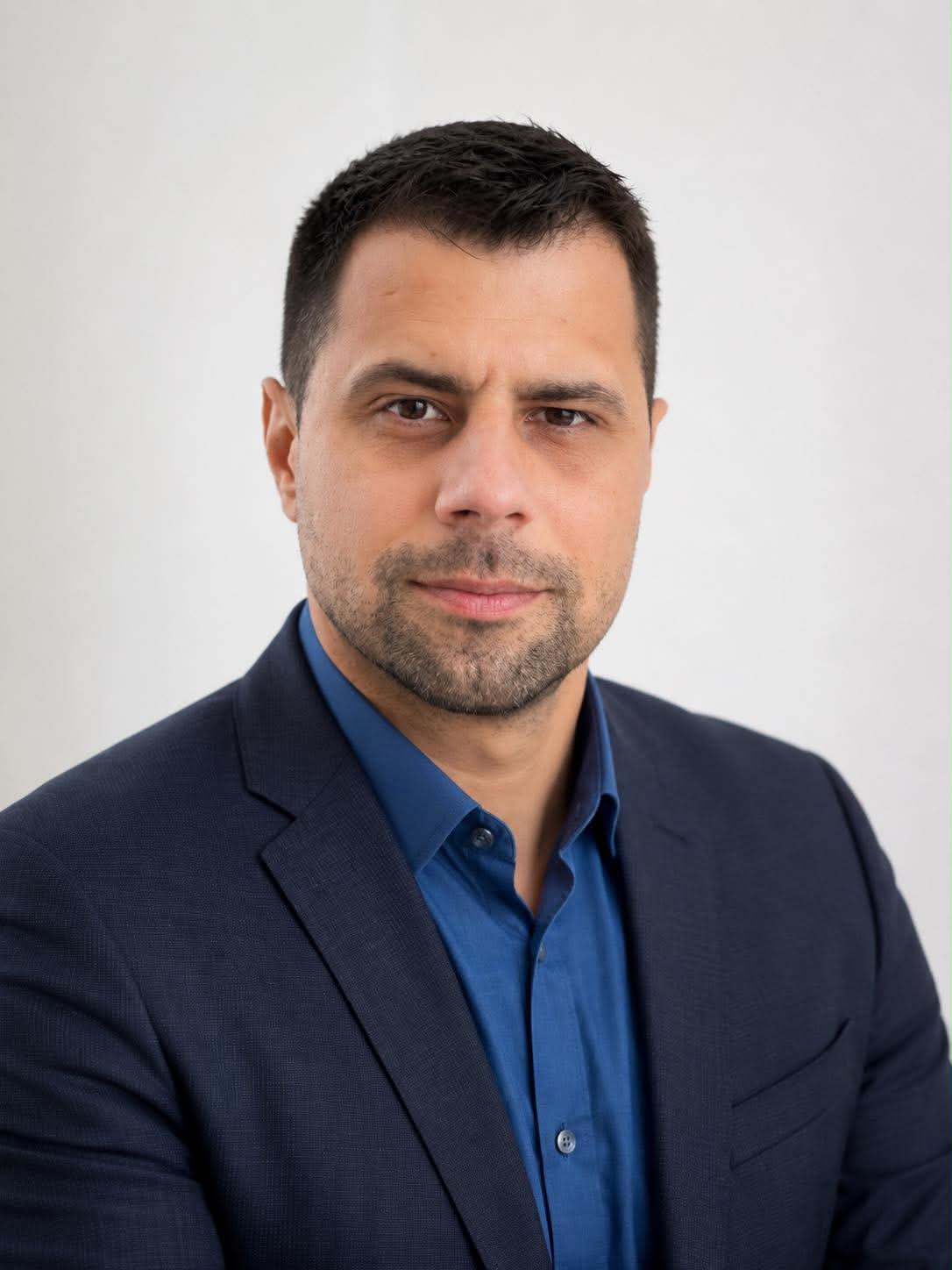}}]{Juraj Gazda}
received the Ph.D.\ degree from the Technical University of Ko\v{s}ice, Slovakia, in 2010. He is currently a Professor with the Department of Computers and Informatics, Technical University of Ko\v{s}ice. His research interests include 5G/6G networks, artificial intelligence, machine learning, and agent-based modelling of complex networks.
\end{IEEEbiography}

\begin{IEEEbiography}[{\includegraphics[width=1in,height=1.25in,clip,keepaspectratio]{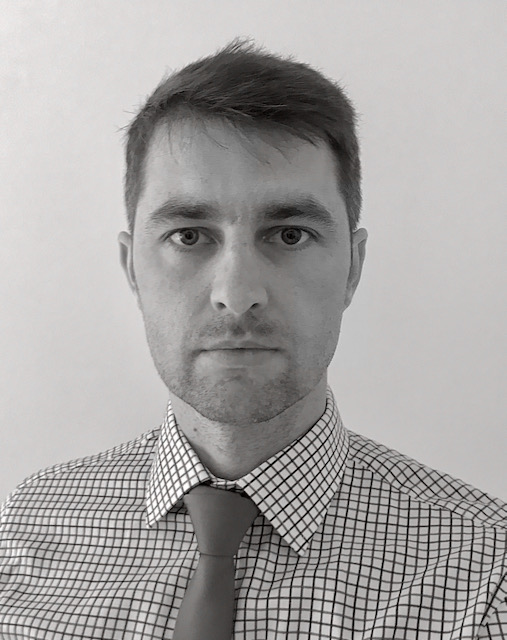}}]{Peter Drot\'ar}
received the M.Sc.\ and Ph.D.\ degrees in electronics from the Technical University of Ko\v{s}ice, Slovakia, in 2007 and 2010, respectively. He is currently an Associate Professor with the Department of Computers and Informatics, Technical University of Ko\v{s}ice, where he leads the Intelligent Information Systems Laboratory. His research interests include machine learning, medical imaging, biomedical signal processing, handwriting analysis, and feature selection. He is a Member of IEEE and EurAI.
\end{IEEEbiography}

\vskip 0pt plus 1fill

\end{document}